\documentclass[10pt,twocolumn,letterpaper]{article}

\usepackage[1]{wacv} 

\usepackage{graphicx}
\usepackage{amsmath}
\usepackage{amssymb}
\usepackage{booktabs}
\usepackage{multirow}
\usepackage{algorithm}
\usepackage[noend]{algpseudocode}

\usepackage[pagebackref,breaklinks,colorlinks]{hyperref}

\usepackage[capitalize]{cleveref}
\crefname{section}{Sec.}{Secs.}
\Crefname{section}{Section}{Sections}
\Crefname{table}{Table}{Tables}
\crefname{table}{Tab.}{Tabs.}

\def\wacvPaperID{*****} 
\def\confName{WACV}
\def\confYear{2024}

\begin{document}

\title{Evaluating Supervision Levels Trade-Offs\\ for Infrared-Based People Counting}

\author{David Latortue\thanks{Equal contribution.}\\
{\tt\small david.latortue.1@ens.etsmtl.ca}
\and
Moetez Kdayem\footnotemark[1]\\
{\tt\small moetez.kdayem.1@ens.etsmtl.ca}
\and
Fidel A. Guerrero Pe\~{n}a\\
{\tt\small fidel-alejandro.guerrero-pena@etsmtl.ca}
\and
Eric Granger\\
{\tt\small eric.granger@etsmtl.ca}
\and
Marco Pedersoli\\
{\tt\small marco.pedersoli@etsmtl.ca}\\
LIVIA, Dept. of Systems Engineering\\
ETS Montreal, Canada\\
}
\maketitle

\begin{abstract}
    Object detection models are commonly used for people counting (and localization) in many applications but require a dataset with costly bounding box annotations for training. Given the importance of privacy in people counting, these models rely more and more on infrared images, making the task even harder. In this paper, we explore how weaker levels of supervision affect the performance of deep person counting architectures for image classification and point-level localization. Our experiments indicate that counting people using a convolutional neural network with image-level annotation achieves a level of accuracy that is competitive with YOLO detectors and point-level localization models yet provides a higher frame rate and a similar amount of model parameters. Our code is available at: https://github.com/tortueTortue/IRPeopleCounting.
\end{abstract}


\section{Introduction}
\label{sec:intro}
Intelligent building systems aim to enhance energy efficiency, environmental sustainability, security, and safety while improving user comfort \cite{HABASH20221}. Various subsystems within the building monitor and control lighting, heating, ventilation and air conditioning (HVAC), and other energy-consuming systems and manage space utilization to promote occupant well-being. These systems may leverage people counting technology to monitor occupancy, optimize energy consumption, and contribute to security and safety. However, privacy is a significant concern when implementing such technology \cite{DBLP:journals/corr/abs-2010-11929}. Standard privacy-preserving measures, including data minimization and infrared cameras, help maintain individual anonymity and ensure privacy in low-light conditions. Therefore, this paper focuses on people counting using images captured with infrared cameras. Additionally, this paper focuses on methods for counting people in sparsely occupied scenes (not dense crowds) with limited overlap among people -- a setting we call sparse crowd counting.

Commonly, the preferred approaches for counting people in sparse settings involve using object detectors \cite{FAN2022224}\cite{DBLP:journals/corr/abs-2007-12831}. However, these approaches require annotating bounding boxes around individuals, which can be time-consuming and expensive. Fortunately, there are more cost-effective methods that can achieve the same goal with a lower cost for annotations.
In particular, point-wise based supervision is often used in crow counting techniques \cite{song2021rethinking,Liu_2023_ICCV}, and relies solely on $(x,y)$ pixel coordinates to localize people. This typically reduces the number of clicks needed for annotation by half, and eliminates the need for adjustments when bounding boxes do not properly fit a person. At the lowest level of supervision, image-level\cite{sun2018fishnet} counting relies on a single integer value annotation per image to indicate the number of people in each image. In this study, the level of supervision refers to the information available during the annotation rather than the learning process.

\begin{figure}
    \begin{tabular}{c}
        \includegraphics[width=.8\linewidth]{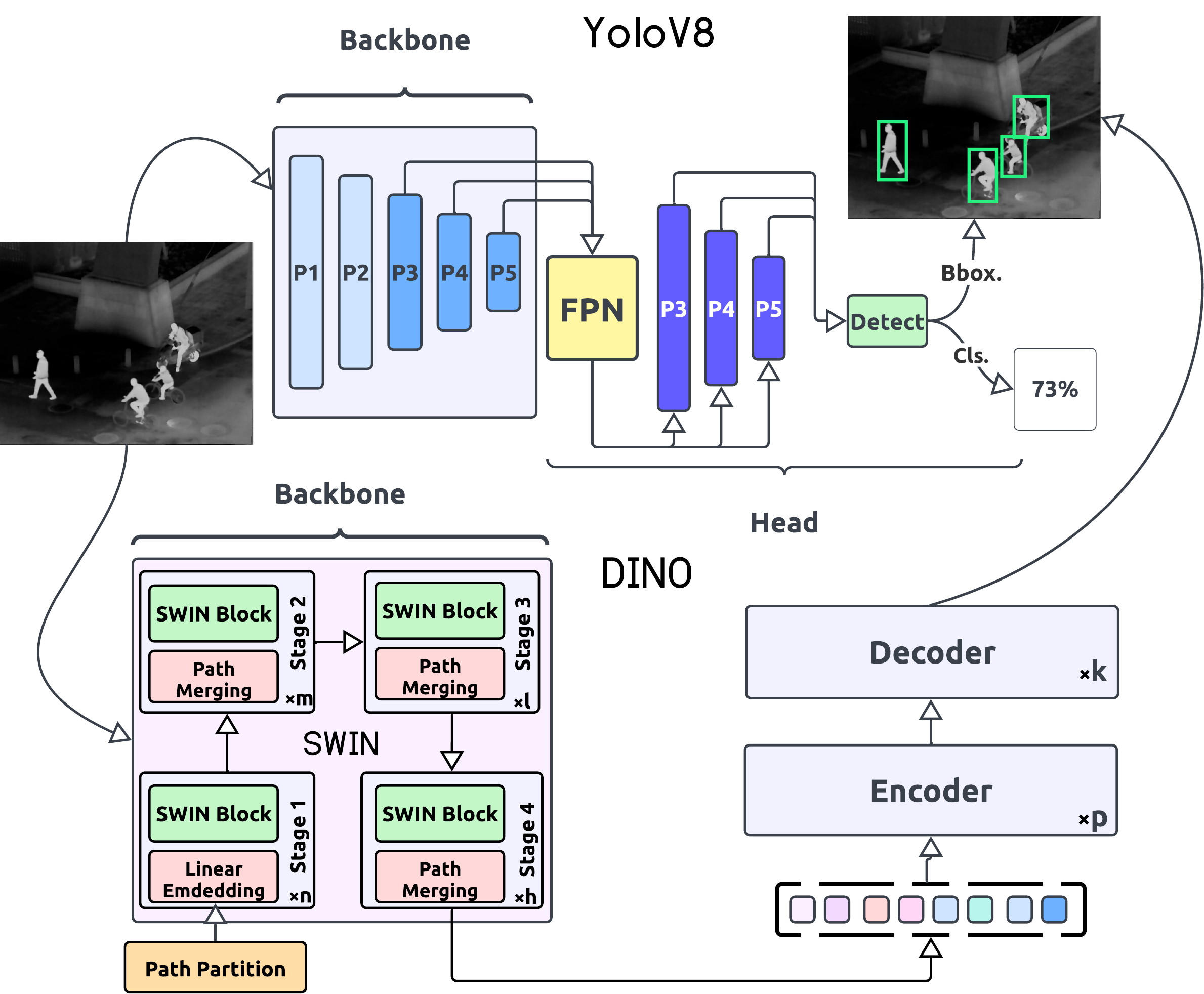} \\
        (a) Object detectors: stronger supervision.                       \\ \\
        \includegraphics[width=.8\linewidth]{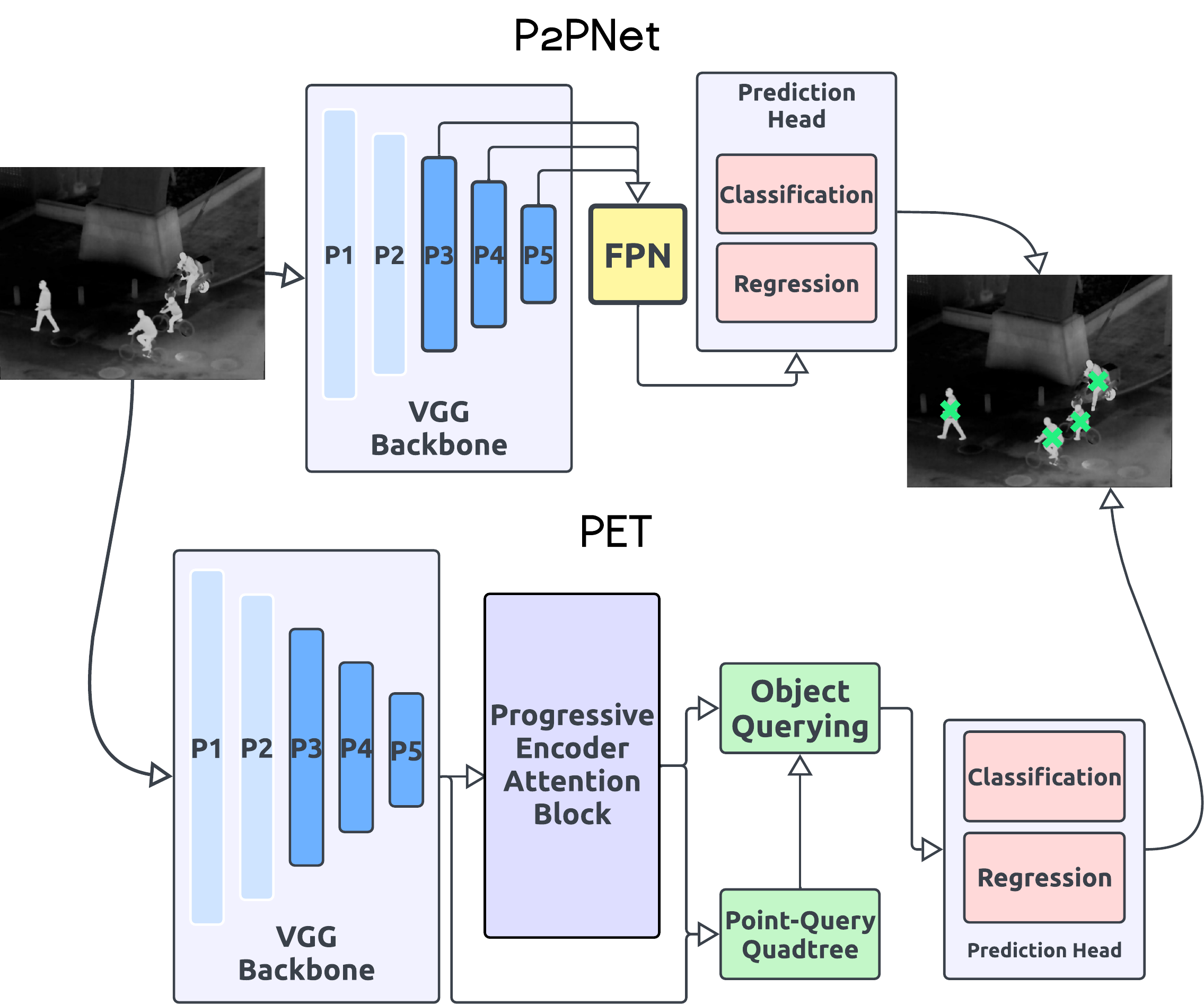}   \\
        (b) Point-wise localization: strong supervision.                  \\ \\
        \includegraphics[width=.8\linewidth]{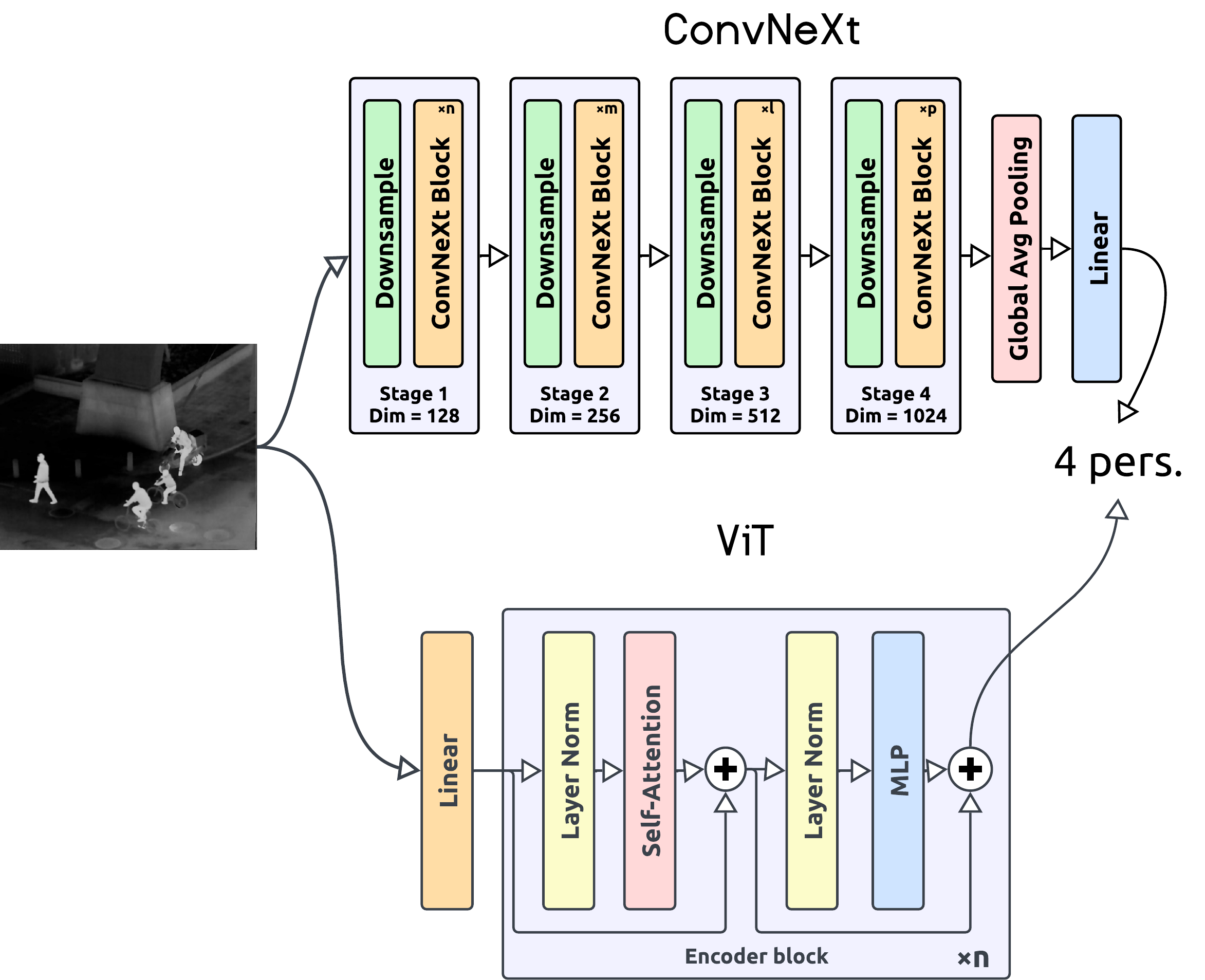}   \\
        (c) Image-level counting: normal supervision.                     \\ \\
    \end{tabular}
    \caption{A dichotomy of the DL architectures analyzed in this paper sorted according to the level of supervision, from least expensive (bottom) to most expensive (top) annotation process.}
    \label{fig:models}
\end{figure}

In this paper, we investigate the impact of the three aforementioned levels of supervision on the accuracy of deep learning (DL) models for sparse people counting based on infrared images. Our study compares the accuracy and complexity of YoloV8 and DINO object detection models, P2PNet and PET point-level localization models, as well as ConvNeXt and ViT image-level counting models. \cref{fig:models} presents a dichotomy of all these models. Additionally, to further improve the performance of the image-level models, we also employ the masked autoencoder (MAE) pretraining method to fully exploit the potential of using unlabeled data. We finally explore the effect of the dataset size and localization results, utilizing the most popular and cost-effective models in each category.

Our main contributions are summarized as follows.
(1) We provide an extensive empirical comparison of various people counting techniques across multiple levels of supervision on two infrared image datasets (LLVIP and Distech IR). Our results show that image-level counting architectures deliver comparable performance to detectors, significantly reducing the annotation effort required.
(2) The MAE pretraining technique is utilized in the people counting task, resulting in improved performance, especially when dealing with large amounts of unannotated data.

\section{Related Work}
\label{sec:formatting}

\textbf{People counting} is a computer vision task that aims to estimate the number of individuals in an image. To address this challenge, various approaches can be employed, ranging from density estimation \cite{DBLP:journals/corr/abs-1908-03314} to object detection \cite{Ren17}. In the literature, these methodologies are categorized as part of the broader group of methods related to crowd counting\cite{FAN2022224}. For very dense crowds, the techniques proposed in the literature tend to prioritize localizing the head of each individual\cite{DBLP:journals/corr/abs-2009-13077,DBLP:journals/corr/abs-1811-10452,Ma_2022} or density map estimation\cite{7533041,7410729,6460719}. In the context of intelligent buildings, sparse crowds are more prevalent. Therefore, our work focuses on people counting for scenarios with 0 to 20 people. For this, DL models for object detection, object localization, and image-level estimations\cite{inbook} are employed.

\textbf{Object detectors} predict the location and class label for each objection, aiming to identify the position and nature of objects in an image\cite{DBLP:journals/corr/abs-2106-11342}. Various DL models have been developed for this task, significantly improving performance in recent years. In the literature, these methods are often categorized as either single-shot detection or two-shot detection. Single-shot (or single-stage) detection is an approach where the model directly outputs the position of an object, its bounding boxes, and the probability of its category. On the other hand, a two-stage model extracts an ensemble of regions of interest (ROIs) and then classifies each.
The most well-known model families for single-shot and two-shot detectors are YOLO \cite{DBLP:journals/corr/RedmonDGF15,DBLP:journals/corr/RedmonF16,DBLP:journals/corr/abs-1804-02767} and R-CNN\cite{DBLP:journals/corr/Girshick15,DBLP:journals/corr/GirshickDDM13,DBLP:journals/corr/RenHG015}, respectively. A later addition to the single-shot detector category is the DETR\cite{10.1007/978-3-030-58452-8_13}, a transformer-based detector that eliminates the need for hand-crafted components.
DETR Improved Denoising Anchor Boxes (DINO)\cite{zhang2023dino}, combines DAB-DETR\cite{liu2022dabdetr} and Denoising-DETR\cite{Li_2022_CVPR},  for improved efficiency. It refines anchor boxes and classifications through a mixed query selection strategy and contrastive denoising training. In this work, for people counting, we use YOLOs, which have proven to be the best compromise between efficiency and performance, and DINO to investigate whether transformers can achieve the performance of earlier proposed architectures in this domain.

\textbf{Point-wise localization} is a task that involves locating an instance of an object within an image. Traditionally, object localization refers to identifying the coordinates of an object as well as defining bounding boxes around the object\cite{DBLP:journals/corr/RhodesQM16}. However, there are models exclusively designed to focus solely on the object's position\cite{song2021rethinking,Liu_2023_ICCV}, which we refer to as point-level localizes. In the literature, numerous approaches are employed for crowd counting using coordinates like SGANet\cite{wang22crowds}, GauNet\cite{Cheng_2022_CVPR}, LoViTCrowd\cite{Tran_2022_BMVC}. However, the two models currently leading in state-of-the-art performance on the popular benchmark ShanghaiTech A\cite{7780439} are PET\cite{Liu_2023_ICCV} and P2PNet\cite{song2021rethinking}. Therefore, we have chosen to use these for our work.

\textbf{Image-level counting}
typically involves categorizing the content of single-object images \cite{DBLP:journals/corr/abs-1912-12162}. In this work, we have utilized the two common DL models for image classification/regression that often provide state-of-the-art performance on benchmark datasets: the CNN and Vision Transformer (ViT) \cite{DBLP:journals/corr/abs-2010-11929}. Furthermore, image-level supervision for people counting is straightforward, typically requiring only the count of people in each image.
However, we can distinguish two image-level approaches for people counting, through classification and regression. For both ViTs and CNNs, there is a wide range of high-performing models to choose from such as SWIN\cite{DBLP:journals/corr/abs-2103-14030}, VOLO\cite{DBLP:journals/corr/abs-2106-13112}, RevCol\cite{cai2023reversible}, NFNet\cite{DBLP:journals/corr/abs-2102-06171}. In this paper, we have selected ConvNeXt and ViT for their efficiency and simplicity.

\section{Methodology}
Our study compares different levels of supervision used to train a DL model for the people counting task.
In particular, we define three levels of supervision based on the strength of annotations used for people counting:\\
\noindent \textbf{1) Normal supervision:} In this level, the full annotation consists of the people count.\\
\noindent \textbf{2) Strong supervision:} This level involves using point-level localization, i.e., pixel coordinates of the person center, as annotations.\\
\noindent \textbf{3) Stronger supervision:} At this level, bounding boxes enclosing a person are used as annotations.

This section details the architectures and training approaches employed for each supervision strategy for people counting. \cref{fig:models} depicts the architectures in each case.

\subsection{ Stronger Annotations: Object Detection}
Let us consider a set of training samples $\mathcal{D}=\left\{ (x_{i},B_{i}) \right\}$ where $x_{i} \in \mathbb{R}^{W\times H \times C }$ are images with spatial resolution $W\times H$ and $C$ channels. Here, a set of bounding boxes is represented by
$B_{i}=\left\{b_0,b_1,\ldots,b_N \right\}$ with $b=\left(c_x,c_y,w,h \right)$ being $c_x$ and $c_y$ the coordinates of the bounding box with size $w\times h$.
For the sake of simplicity, we omitted the class label of each bounding box from the notation, as the people counting task is concerned exclusively with the person class.
Then, in the training process of a neural network-based detector, we aim to learn a parameterized function $f_{\theta}\colon \mathbb{R}^{W\times H\times C} \to \mathcal{B}$, being $\mathcal{B}$ the family of sets $B_i$ and $\theta$ the parameters vector. For such, the optimization is guided by a loss function, which is a combination of a regression $\mathcal{L}_{reg}$ and a classification $\mathcal{L}_{cls}$ term, i.e., $l_2$ loss and binary cross-entropy, respectively. Averaging these values for every detected instance and sample in $\mathcal{D}$ will yield a cost function that is a surrogate for the task's objective, counting the number of people $y$.
\begin{equation}
    \label{eqn:det}
    \mathcal{C}_{det}(\theta)=\frac{1}{|\mathcal{D}|}\sum_{(x,B)\in\mathcal{D}}\mathcal{L}_{cls}(f(x;\theta), B)+ \lambda \mathcal{L}_{reg}(f(x;\theta), B)
\end{equation}
Then, in the context of people counting, the number of people for a given input image is obtained as the number of valid bounding boxes within the network's output.

This study utilizes two kinds of detectors: YOLO and DINO.  For YOLO, we use version 8, which is the latest model in this family\cite{Ren17}\cite{DBLP:journals/corr/RedmonDGF15}\cite{DBLP:journals/corr/RedmonF16}\cite{DBLP:journals/corr/abs-1804-02767}. Such an architecture is divided into two main components: the backbone and the head. The backbone is a convolutional network that extracts feature maps from the input image. This process generates three scales from the feature maps $P_{i\in(3,5)}$. For each feature map $P_i$, the remaining feature maps $P_j$ and $P_k$ are either upsample using a bilinear interpolation or downsampled via a convolution layer before concatenation. Finally, a detection block outputs bounding boxes and a class prediction for each resulting feature map.

On the other hand, our second choice of detector is DINO, from the DETR family. It is a fully transformer-based model using SWIN\cite{Liu_2021_ICCV} as a backbone. The SWIN transformer backbone is similar to ViT but uses different window sizes at every stage to learn different resolutions. The features yielded by each stage are sent as inputs to the encoder of the DINO detector. This detector is an end-to-end architecture using encoder and decoder blocks coupled with multiple prediction heads to output the prediction boxes.

\subsection{Strong Annotations: Point-Level Localization}

We employ point-level localization as the second level of supervision. Similar to the previous approach, we define a training set $\mathcal{S}=\left\{(x_i, P_i)\right\}$ being $x_i$ the same input images. However, unlike the previous level, the annotation set $P_i$ consists of a collection of pixel coordinates that identify a person's location, along with its confidence score, denoted as $p=(c_x,c_y, s), \forall p\in P_i$. The expected mapping takes the form of $g_{\vartheta}\colon \mathbb{R}^{W\times H\times C} \to \mathcal{P}$ where $\mathcal{P}$ represents the family of possible outputs $P$. Then, the task of finding the optimal parameters vector $\vartheta$ is solved using a Gradient Descent-based optimization that uses the cost function:
\begin{equation}
    \label{eqn:loc}
    \mathcal{C}_{loc}(\theta)=\frac{1}{|\mathcal{S}|}\sum_{(x,P)\in\mathcal{S}}\mathcal{L}_{cls}(g(x;\vartheta), P)+ \lambda \mathcal{L}_{loc}(g(x;\vartheta), P)
\end{equation}
Here, $\mathcal{L}_{cls}$ is the binary cross entropy loss function, and $\mathcal{L}_{loc}$ is the $l_2$ loss between the estimated and the ground truth points. Following the previous setup, the number of people is determined as the number of estimated points with a confidence score higher than a given threshold.

Regarding point-level architectures, we employ P2P-Net\cite{song2021rethinking} and Point query Transformer (PET)\cite{Liu_2023_ICCV}. In the first approach, the network incorporates the first 13 convolutional layers from VGG-16 to extract deep features. These features are subsequently upsampled by a factor of two using nearest-neighbor interpolation and combined with a feature map from a lateral connection via element-wise addition. This lateral connection reduces the channel dimensions of the feature map after the fourth convolutional block. The merged feature map then undergoes a $3 \times 3$ convolutional layer, which helps mitigate the aliasing effect caused by the upsampling process. Finally, a prediction head with two branches generates point locations and confidence scores. Here, we use the same architecture for both branches, consisting of three stacked convolutional layers with interleaved ReLU activations.

On the other hand, the PET network employs a decomposable query process by dividing sparse points into new points and selectively querying them, particularly in densely populated areas. The framework comprises two main components: a point query quadtree and a progressive rectangle attention mechanism. It begins with input passed through a VGG16-based feature extraction backbone \cite{DBLP:journals/corr/SimonyanZ14a}, allowing for scalable quantity estimation. A quadtree, transitioning from sparse to dense, splits each point into 4 points, adaptively partitioning points in crowded scenes. A CNN function encodes pixel localization as a point query, and a transformer decoder decodes the point queries before passing them through a prediction head to obtain crowd predictions. The framework supports different query numbers per image. It employs a progressive pooling approach based on horizontal windows due to its tendency to contain more people than a vertical window, adapting window size according to point density. The decoder's attention focuses on local windows to determine whether a point query corresponds to a person based on context. This approach improves computational efficiency while addressing the crowd-counting challenge.

\subsection{Normal Annotations: Image-Level Counting}

At the lowest level of supervision, we focus on image-level tasks. This task operates on a training dataset, denoted as $\mathcal{T}=\{(x_i, y_i)\}$, containing pairs of images, $x_i$, and the respective count of people present in each image, $y_i$. The parameterized mapping is defined as $h_\phi \colon \mathbb{R}^{W \times H \times C} \to \mathcal{Y}$, where $\mathcal{Y}={0, 1, 2, \ldots, 20}$. The optimization problem at this level aims to minimize a classification cost using the cross-entropy loss function or a regression cost with the $l_2$ loss. The estimated number of people corresponds to the appropriate class in the classification scenario.

At this level of annotation, the first architecture we consider is the Vision Transformer (ViT)\cite{DBLP:journals/corr/abs-2010-11929}. The input images are divided into patches and injected with positional embeddings, later fed to the network. ViT consists of multiple encoder blocks arranged in sequence, followed by a classification head represented by a simple multi-layer perceptron. Each encoder block comprises a Self-Attention layer and a residual connection to a feedforward network. Such a Self-Attention layer's primary purpose is to model the relationship between different features and prioritize the most relevant ones. This is achieved by having the layer learn a representation for each feature, allowing it to compute the attention score --score of importance-- for every pair of features using dot products of their respective vectors. The goal is to enable the model to capture long and short-range relationships between features. One drawback of ViT is its requirement for a substantial amount of data to converge. Hence, we chose to experiment with a pretraining approach.

Despite the success of ViT-based methods, the Convolutional Neural Networks have been the default deep learning model for vision-related tasks for a long time. Their success can be attributed to the induced bias introduced by the convolution layers, which encourages the model to learn short-range feature maps. This strong prior has helped the field of vision to reach several success stories in the last decade. The powerful track record of this model and the recent advances with the transformer have inspired researchers to build a more modern version of the CNN following the new architecture designs often used for Transformers. ConvNeXt\cite{DBLP:journals/corr/abs-2201-03545}, a member of the ResNets\cite{DBLP:journals/corr/HeZRS15} and ResNexts\cite{DBLP:journals/corr/XieGDTH16} model families builds on the concept of residual layers and the aggregation of multiple transformation paths within a block. Like its predecessors, ConvNeXt is divided into four stages, where at the entrance of each, the inputs are down-sampled before going through a series of depthwise convolutions. The full overview of the ConvNeXt architecture can be viewed at \cref{fig:models}.
\vspace{-.2cm}

\subsubsection{Masked Autoencoder Pretraining}
Unsupervised pretraining is a common practice in image classification. One promising technique employed for both ViT and ConvNeXt is the Masked Autoencoder \cite{DBLP:journals/corr/abs-2111-06377}\cite{woo2023convnext} pretraining method. It is an unsupervised training objective to reconstruct a full image from a small, random portion of it. This strategy helps image classifiers learn more robust representations to improve classification accuracy.

Similar to standard Autoencoders, this method divides the model into two components: an encoder and a decoder. During pretraining, the classifier functions as the encoder, and the decoder is discarded during fine-tuning. In the training process, the encoder takes a random portion of the image as input, applies a mask to complete the missing parts, and then sends the output to the decoder to generate the full image. The mean squared error loss is computed by comparing the generated image with the original full image. Since this pretraining method was initially developed for image classification, its effectiveness was uncertain for people counting.

\section{Results and Discussion}
\label{sec:results}

\begin{figure*}[t]
    \centering
    \begin{tabular}{cc}
        \begin{subfigure}{0.5\linewidth}
            \centering
            \includegraphics[width=1.5in,height=1.2in]{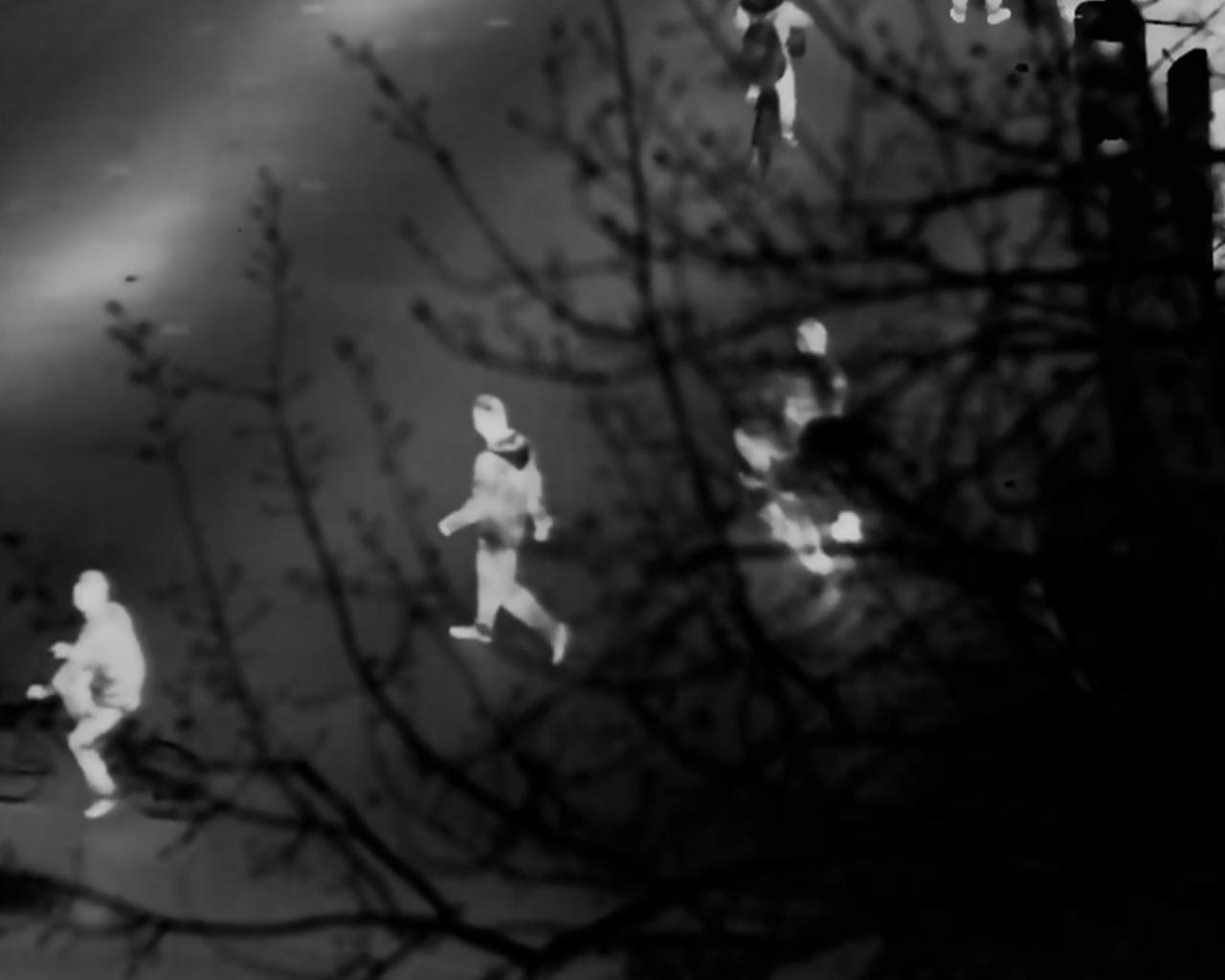}
            \includegraphics[width=1.5in,height=1.2in]{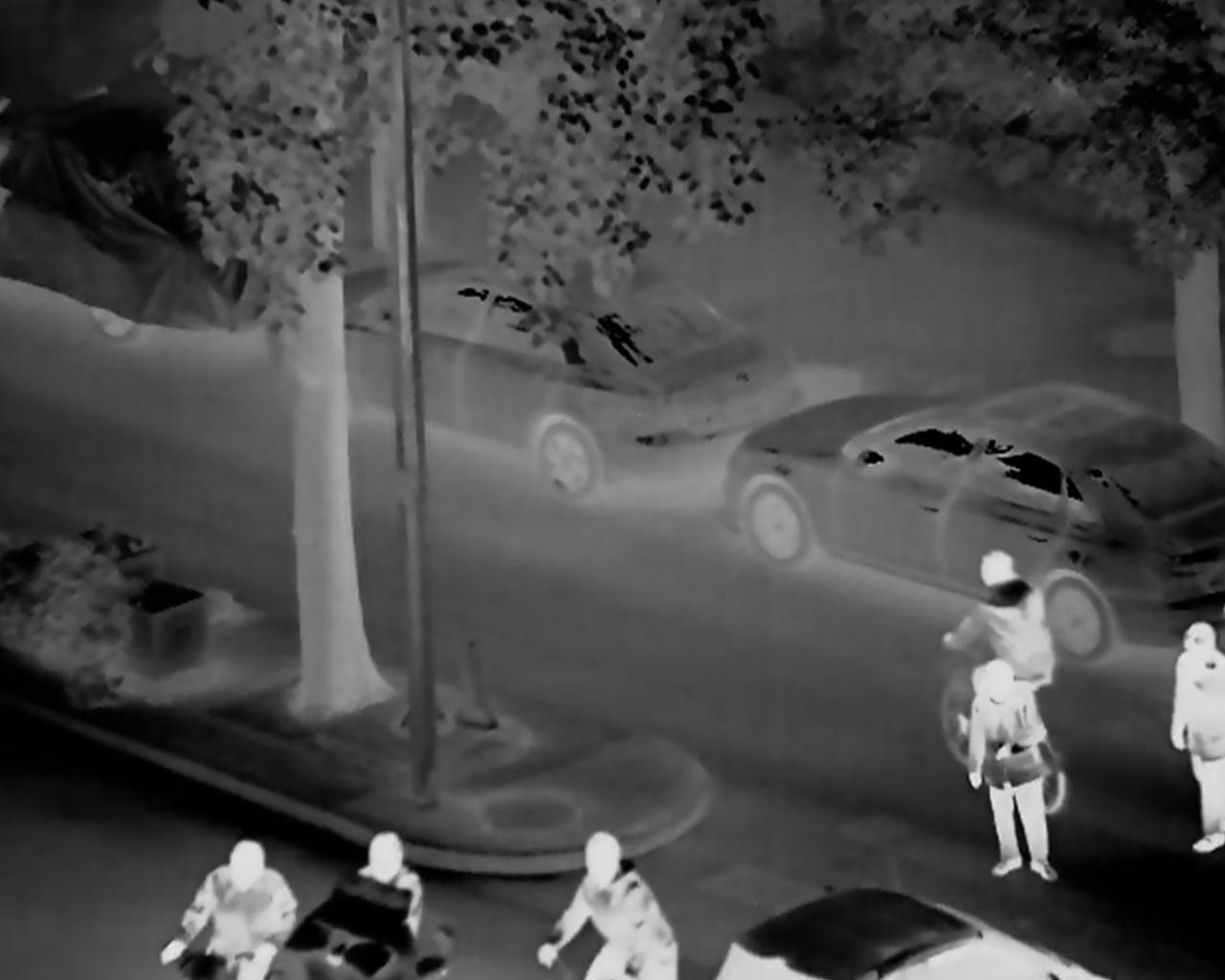}
            \caption{LLVIP}
            \label{fig:dataset-a}
        \end{subfigure} &
        \begin{subfigure}{0.5\linewidth}
            \centering
            \includegraphics[width=1.5in,height=1.2in]{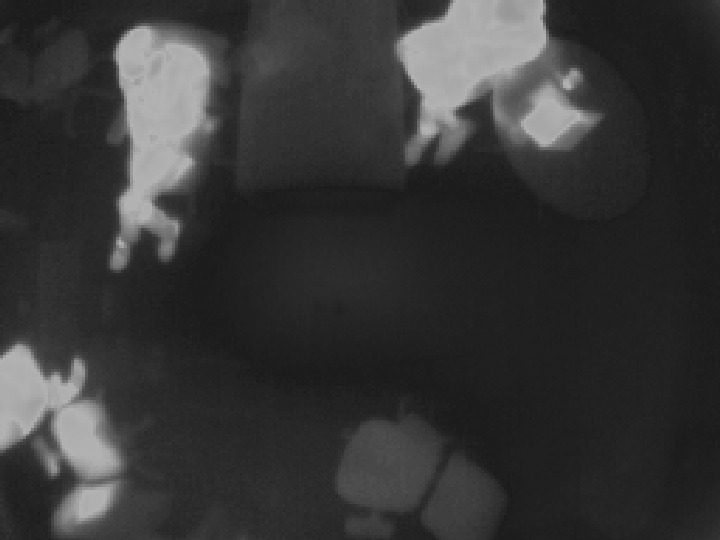}
            \includegraphics[width=1.5in,height=1.2in]{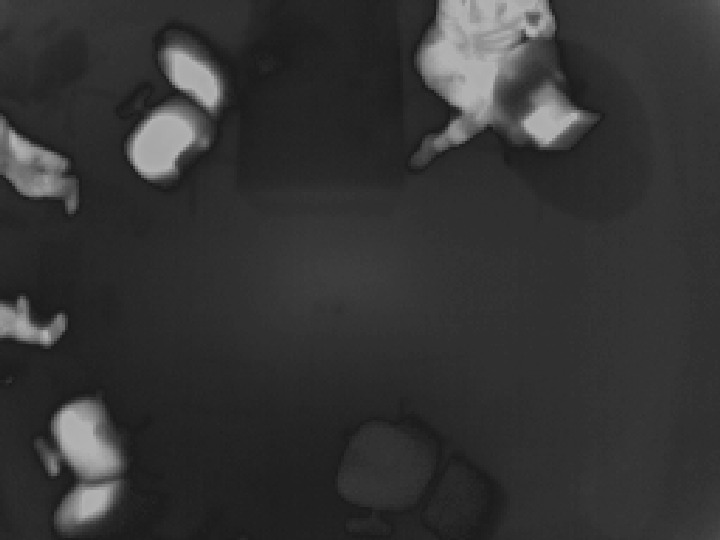}
            \caption{Distech IR}
            \label{fig:dataset-b}
        \end{subfigure}
        \\
    \end{tabular}
    \caption{Examples of infrared images taken respectively from the Distech IR and LLVIP datasets.}
    \label{fig:dataset}
\end{figure*}

\subsection{Experimental Methodology}

\noindent \textbf{1) Datasets:}
In our study, we employed two IR datasets. The first, LLVIP\cite{DBLP:journals/corr/abs-2108-10831}, comprises 15488 pairs of both RGB and IR versions of the same images. Although LLVIP offers both modalities, our focus in this work centers exclusively on IR video surveillance scenarios. As a result, we only used the IR images from the dataset. These IR images are captured by fixed outdoor cameras, providing frames from various street settings and with diverse perspectives.
It's worth noting that the camera perspectives in the original training and testing sets differ. We merged and then re-split the training and testing sets to address the out-of-distribution scenario caused by the differences in camera perspectives between both sets. The dataset is split into 12025 images for training and 3463 images for testing. Each image has annotations consisting of bounding boxes enclosing each person, and the number of people per image can range from 0 to 13. Some examples of the images in the dataset can be seen in \cref{fig:dataset-a}.

The second dataset is the Distech IR\cite{dubail2022privacypreserving}. This dataset consists of IR images capturing individuals in indoor settings across ten office rooms. It encompasses 2536 images, divided into their subsets: 1798 images in the training set, 483 images in the validation set, and 255 images in the testing set. The cameras are positioned statically, providing a top-view perspective. In addition, the dataset contains 2536 sequences, each comprising 64 frames, with only one frame annotated in each sequence. Each image is annotated with bounding boxes around a person. Unlike LLVIP, these frames only include temperature values. To reduce the effect of abnormal values, we applied Winsorization\cite{hastings1947low}, a normalization technique that trims outlier values below 5\%  of the standard deviation and above 95\% of the standard deviation. Distech IR images contain some people ranging between 0 and 12. \cref{fig:dataset-b} displays examples of office scenarios captured with Distech's cameras.

\noindent \textbf{2) Experimental setup:}

\textit{Image-level counters:}
Every image-level counting method in this study was trained using a consistent pipeline. We trained all models for 400 epochs using the AdamW optimizer and a batch size of 64. All additional training details specific to each model variation are provided in supplementary material. These parameters remained uniform for both LLVIP and Distech IR datasets. We used the ConvNeXt-Tiny and ConvNeXt-Micro architectures for both classification and regression. Additionally, our experiments involved the ViT-4L and ViT-3L networks. All models used the same data split to ensure a fair benchmarking process. We rounded the output to the nearest integer to determine the people count in the regression-based approaches. We applied the maximum-a-posteriori decision rule in classification-based approaches, matching the people count with the corresponding class.

    {\it Point-wise localization:}
For both P2P-Net and PET, each image undergoes a feature extraction process using a pre-trained VGG-16 backbone. The optimization is carried out using Adam optimizer \cite{loshchilov2018decoupled} with a weight decay of $5 \times 10^{-4}$. P2P-Net, trained for 3500 epochs as suggested by its authors, utilizes specific parameters critical to its operation, including a window stride of 8, 4 reference points, and a backbone learning rate of $10^{-5}$. The batch size for both models was set to $8$. In the case of PET, which is typically trained for 1500 epochs, the CNN backbone (VGG16) and the transformer have learning rates of $10^{-5}$ and $10^{-4}$, respectively. The point-query quadtree has a maximum depth of 2, and the initial sparse point query stride is $K = 8$. The transformer encoder and decoder layers share a common quadtree decoder. Window parameters are $s_e = 16$ and $r_e = 2$. Loss coefficients are $\lambda_1 = 5.0$ and $\lambda_2 = 0.1$, which help balance the loss function terms.

    {\it Object detectors:}
YoloV8 post-processing relies on two hyper-parameters: the confidence and non-maximal suppression thresholds, both within the range of 0 to 1. We tuned these thresholds for all detectors by selecting the values that resulted in the highest count accuracy on the validation set. Our experiments use YoloV8-L, YoloV8-M, and YoloV8-S architectures. The specific values chosen for each network are in the supplementary materials. For DINO, we train it for 12 epochs with a learning rate of $10^{-4}$, a batch size of 4, and a weight decay of $10^{-4}$ using the SWIN-TINY backbone. The confidence threshold for DINO was tuned following the same procedure as before.

\subsection{Main Comparative Results}
In this section, we compare the three levels of supervision for people counting on LLVIP and Distech, and then the three levels of supervision on LLVIP with different amounts of training data. In addition, we analyze the impact of the counting model normal annotations of regression and classification heads, and the use of a masked autoencoder pre-training. Finally, we compare the computational and localization performance of methods.

\noindent \textbf{1) Counting with different levels of supervision: }
\begin{table}[t]\centering
    \caption{Count accuracy results on LLVIP}\label{tab:accllvip2}
    \footnotesize
    \begin{tabular}{lrrrr}\toprule
        \textbf{Model} & \textbf{Acc↑}     & \textbf{MSE↓}  & \textbf{MAE↓}  \\\midrule
        YoloV8-L       & \textbf{87.86 \%} & 0.191          & 0.160          \\
        YoloV8-M       & 87.80 \%          & 0.182          & 0.156          \\
        YoloV8-S       & 86.36 \%          & 0.215          & 0.180          \\
        DINO           & 87.38 \%          & \textbf{0.150} & \textbf{0.140} \\\cmidrule{1-4}
        P2PNet         & 56.22 \%          & 0.955          & 0.578          \\
        PET            & 59.19 \%          & 0.776          & 0.515          \\\cmidrule{1-4}
        ConvNeXt-Tiny  & 80.13 \%          & 0.239          & 0.211          \\
        ViT-4L         & 63.89 \%          & 0.446          & 0.383          \\
        ConvNeXt-Micro & 80.59 \%          & 0.227          & 0.204          \\
        ViT-3L         & 61.29 \%          & 0.491          & 0.421          \\
        \bottomrule
    \end{tabular}
\end{table}
Tables \ref{tab:accllvip2} and \ref{tab:accdistech2} present count accuracy, mean squared error, and mean absolute error results for each model on LLVIP and Distech IR, respectively. Based on count accuracy results, modern image-level approaches are beginning to narrow the performance gap with object detection techniques, even surpassing some state-of-the-art crowd counting methods such as P2PNet~\cite{song2021rethinking} and PET~\cite{Liu_2023_ICCV}. In the case of Distech IR, the ConvNeXt-Micro image classifier with Masked Autoencoder pretraining achieved highly competitive results across all YoloV8 sizes. For LLVIP, both ViT and ConvNeXt-based classifiers outperformed P2PNet and PET. However, it's important to note that, in LLVIP, the best classifier, ConvNeXt-Micro, achieved a count accuracy of 80.59\%, while the best detector, YoloV8-L, reached 87.86\%. Depending on the application and labeling budget, this may represent a reasonable trade-off. It is also worth noting that, despite not having the best accuracy, DINO still manages to have the lowest MSE and MAE for both datasets. This could mean that DINO's mistakes are closer to the correct count than the other models. Another interesting observation is that, despite having stronger supervision, localization models do not outperform the classifiers on LLVIP. This might be attributed to the fact that these models were initially designed to handle denser crowds.

\begin{table}[t]\centering
    \caption{Count accuracy results on Distech IR}\label{tab:accdistech2}
    \footnotesize
    \begin{tabular}{lrrrr}\toprule
        \textbf{Model} & \textbf{Acc↑}     & \textbf{MSE↓}  & \textbf{MAE↓}   \\\midrule
        YoloV8-L       & 90.02 \%          & 0.173          & 0.123           \\
        YoloV8-M       & 88.15 \%          & 0.154          & 0.133           \\
        YoloV8-S       & \textbf{91.06} \% & 0.185          & 0.123           \\
        DINO           & 90.98 \%          & \textbf{0.114} & \textbf{0.098 } \\\cmidrule{1-4}
        P2PNet         & 76.47\%           & 0.271          & 0.247           \\
        PET            & 84.70\%           & 0.282          & 0.196           \\\cmidrule{1-4}
        ConvNeXt-Tiny  & 88.24 \%          & 0.721          & 0.266           \\
        ViT-4L         & 79.61 \%          & 1.323          & 0.485           \\
        ConvNeXt-Micro & 88.53 \%          & 0.790          & 0.266           \\
        ViT-3L         & 78.04 \%          & 1.218          & 0.442           \\
        \bottomrule
    \end{tabular}
\end{table}

\begin{figure}
    \includegraphics[width=\linewidth]{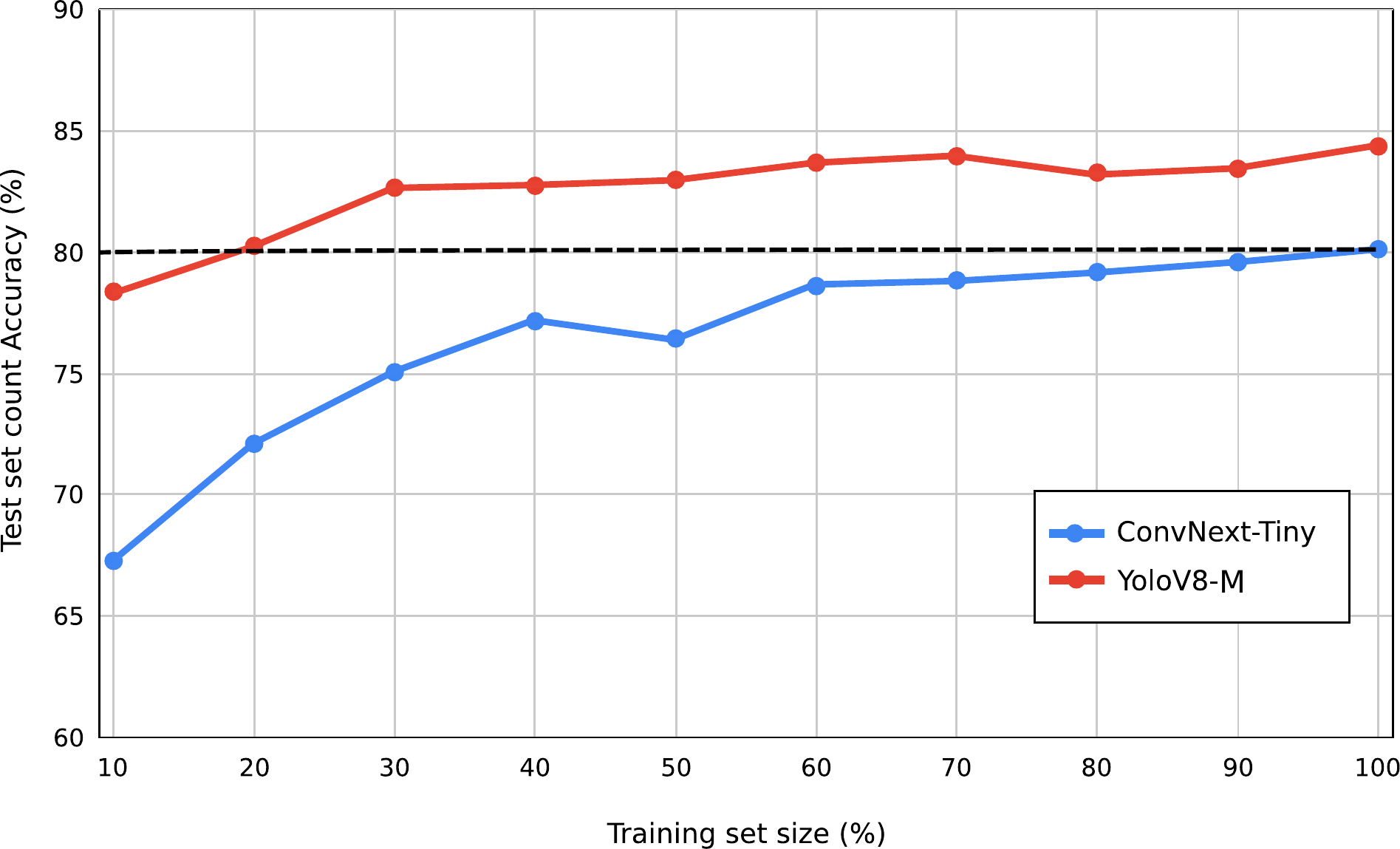}
    \caption{Count accuracy on the test set depending on the training set size for LLVIP}
    \label{fig:datasize}
\end{figure}

\noindent \textbf{2) Counting with different amounts of data: }
Based on the count accuracy results in LLVIP, it is evident that with even the proper training regimen, ConvNeXt-Tiny can't entirely close the performance gap with a model trained with stronger supervision like YoloV8-L for the same amount of data. Our results show that ConvNeXt-Tiny achieved an 80.13 $\%$ count accuracy, whereas YoloV8-L achieves 87.86 $\%$. Considering the cost difference of labeling, we believe that those performances nonetheless establish ConvNeXt-Tiny as a potentially viable solution for specific applications. However, we were keen to understand how much training data ConvNeXt would need to reach the performance of YoloV8. To gain a better perspective, we trained both models using various portions of the training data, ranging from 10 $\%$ to 100 $\%$ of LLVIP with 10 $\%$ increments. This experiment allowed us to find a relationship between normal and stronger annotations. We
found that, in practice, YoloV8-L and YoloV8-M required respectively an approximately 16 $\%$ and 17 $\%$ of the amount of labeled data used for ConvNeXt-Tiny to achieve equivalent performance. On the other hand, as mentioned earlier in this paper, bounding box annotations are considerably more time-consuming and challenging to produce than image-level annotations, which involve specifying the number of people in the image. A graph displaying count accuracy on the testing set as a function of the amount of data can be seen in \cref{fig:datasize}.

\noindent \textbf{3) Ablations on image-level counting:}
\cref{tab:accllvip} and \ref{tab:accdistechir} show the obtained result over LLVIP and Distech IR datasets for image-level settings. Let's start by looking at the performances of both architectures: ViT and ConvNeXt. Contrary to our expectations, considering the impressive performances of Transformers on challenging benchmarks like ImageNet, the ConvNeXt significantly outperforms ViT by a substantial margin of at least 14\% in LLVIP and 3\% in the case of Distech IR. This outcome can be attributed to two key factors. First, we utilized the full ConvNeXt model pre-trained on ImageNet, providing a competitive parameter count advantage compared to using only the first four layers of ViT-S. Second, the Transformer, as indicated in the literature\cite{lu2022bridging}, tends to excel with large and diverse datasets, which is not the case in our settings.
Following with using pretraining to improve accuracy, our data shows that the masked auto-encoding method consistently yields better results for all classifier configurations, suggesting that MAE pretraining is also a suitable approach for people counting.

As discussed, we also aim to determine the optimal modeling approach -- classification or regression -- for image-level counting. Our results show that, in the case of LLVIP, regression consistently outperforms classification, regardless of whether pretraining is used. However, the opposite holds for Distech IR, where classification outperforms regression. Upon analyzing the class distribution of both datasets, we observed that the LLVIP dataset exhibits a significantly more balanced distribution, with a standard deviation of 9.86\% across class occurrences, as opposed to 16.03\% for the other dataset.

\begin{table}[t]\centering
    \caption{Count accuracy for different configurations of ConNext and ViT for LLVIP.}\label{tab:accllvip}
    \footnotesize
    \setlength{\tabcolsep}{4pt}
    \begin{tabular}{lccrrrr}\toprule
        \textbf{Model}                  & \textbf{Pretrain}     & \textbf{Head} & \textbf{Acc↑} & \textbf{MSE↓} & \textbf{MAE↓} \\\midrule
        \multirow{4}{*}{ConvNeXt-Tiny}  & \multirow{2}{*}{None} & Class.        & 78.74 \%      & 0.251         & 0.224         \\
                                        &                       & Regr.         & 79.84 \%      & 0.238         & 0.213         \\\cmidrule{2-6}
                                        & \multirow{2}{*}{MAE}  & Class.        & 79 .69 \%     & 0.249         & 0.215         \\
                                        &                       & Regr.         & 80.13 \%      & 0.239         & 0.211         \\\midrule
        \multirow{4}{*}{ViT-4L}         & \multirow{2}{*}{None} & Class.        & 62.36 \%      & 0.480         & 0.408         \\
                                        &                       & Regr.         & 62.94 \%      & 0.478         & 0.403         \\\cmidrule{2-6}
                                        & \multirow{2}{*}{MAE}  & Class.        & 63.20 \%      & 0.505         & 0.409         \\
                                        &                       & Regr.         & 63.89 \%      & 0.446         & 0.383         \\\midrule\midrule
        \multirow{4}{*}{ConvNeXt-Micro} & \multirow{2}{*}{None} & Class.        & 77.99 \%      & 0.264         & 0.233         \\
                                        &                       & Regr.         & 79.03 \%      & 0.253         & 0.223         \\\cmidrule{2-6}
                                        & \multirow{2}{*}{MAE}  & Class.        & 78.14 \%      & 0.274         & 0.235         \\
                                        &                       & Regr.         & 80.59 \%      & 0.227         & 0.204         \\\midrule
        \multirow{4}{*}{ViT-3L}         & \multirow{2}{*}{None} & Class.        & 58.15 \%      & 0.614         & 0.479         \\
                                        &                       & Regr.         & 61.06 \%      & 0.496         & 0.420         \\\cmidrule{2-6}
                                        & \multirow{2}{*}{MAE}  & Class.        & 61.70 \%      & 0.515         & 0.424         \\
                                        &                       & Regr.         & 61.29 \%      & 0.491         & 0.421         \\
        \bottomrule
    \end{tabular}
\end{table}

\begin{table}[t]\centering
    \caption{Count accuracy for different configurations of ConNext and ViT for Distech IR.}\label{tab:accdistechir}
    \footnotesize
    \setlength{\tabcolsep}{3.5pt}
    \begin{tabular}{lccrrrr}\toprule
        \textbf{Model}                  & \textbf{Pretrain}     & \textbf{Head} & \textbf{Acc↑} & \textbf{MSE↓} & \textbf{MAE↓} \\\midrule
        \multirow{4}{*}{ConvNeXt-Tiny}  & \multirow{2}{*}{None} & Class.        & 82.75 \%      & 0.787         & 0.294         \\
                                        &                       & Regr.         & 82.35 \%      & 0.771         & 0.325         \\\cmidrule{2-6}
                                        & \multirow{2}{*}{MAE}  & Class.        & 88.24 \%      & 0.721         & 0.266         \\
                                        &                       & Regr.         & 86.28 \%      & 0.697         & 0.274         \\\midrule
        \multirow{4}{*}{ViT-4L}         & \multirow{2}{*}{None} & Class.        & 79.22 \%      & 1.072         & 0.407         \\
                                        &                       & Regr.         & 76.08 \%      & 1.556         & 0.545         \\\cmidrule{2-6}
                                        & \multirow{2}{*}{MAE}  & Class.        & 79.61 \%      & 1.323         & 0.485         \\
                                        &                       & Regr.         & 67.06 \%      & 1.429         & 0.607         \\\midrule\midrule
        \multirow{4}{*}{ConvNeXt-Micro} & \multirow{2}{*}{None} & Class.        & 85.10 \%      & 0.791         & 0.290         \\
                                        &                       & Regr.         & 78.82 \%      & 0.940         & 0.392         \\\cmidrule{2-6}
                                        & \multirow{2}{*}{MAE}  & Class.        & 88.24 \%      & 0.790         & 0.266         \\
                                        &                       & Regr.         & 86.67 \%      & 0.678         & 0.254         \\\midrule
        \multirow{4}{*}{ViT-3L}         & \multirow{2}{*}{None} & Class.        & 83.53 \%      & 0.936         & 0.341         \\
                                        &                       & Regr.         & 70.98 \%      & 1.400         & 0.545         \\\cmidrule{2-6}
                                        & \multirow{2}{*}{MAE}  & Class.        & 78.04 \%      & 1.218         & 0.442         \\
                                        &                       & Regr.         & 67.84 \%      & 1.194         & 0.529         \\
        \bottomrule
    \end{tabular}
\end{table}

\subsection{Models size and computational cost}
The inference time required for each model becomes crucial when selecting the right model for a solution. While the parameter count can indicate the expected speed performance among models within the same family (e.g., YoloV8), it may not effectively characterize the speed of a model relative to another model type with a similar parameter count. We conducted a frame-per-second benchmark on GPU and CPU for every model to facilitate a fair comparison of models across different types. All benchmarks were performed under the same conditions. The GPU used for benchmarking is an NVIDIA A100 SXM4 with 40GB of memory, and the CPU is AMD EPYC 7413 24-core Processor. Each benchmark started with 100 warm-up steps, followed by 10000 inferences with a batch size of 1. We report the mean of all repetitions. The results can be viewed at \cref{tab:efficiency}. As observed in the table, image-level counting achieves the highest efficiency on the CPU, as expected. This is a desirable feature for intelligent building applications that require close-to-real-time responses.

\begin{table}[!htp]\centering
    \caption{Number of parameters and frames-per-second (FPS) on GPU and CPU for all evaluated models.}\label{tab:efficiency}
    \footnotesize
    \begin{tabular}{lccc}\toprule
        \textbf{Model} & \textbf{Parameters} & \textbf{FPS (GPU)} & \textbf{FPS (CPU)} \\\midrule
        YoloV8-L       & 44 M                & 103.16             & 14.07              \\
        YoloV8-M       & 26 M                & 125.33             & 17.21              \\
        YoloV8-S       & 11 M                & 162.5              & 28.46              \\
        DINO-SWIN-T    & 48 M                & 67.72              & 22.17              \\\cmidrule{1-4}
        P2PNet         & 22 M                & 365.41             & 27.5               \\
        PET            & 21 M                & 70.38              & 9.32               \\\cmidrule{1-4}
        ConvNeXt-Tiny  & 29 M                & 127.92             & 30.45              \\
        ConvNeXt-Micro & 24 M                & 160.35             & 32.84              \\
        ViT-4L         & 30 M                & 316.3              & 51.88              \\
        ViT-3L         & 23 M                & 404.54             & 60.21              \\
        \bottomrule
    \end{tabular}
\end{table}

\subsection{Localization}
Additionally, the counting capabilities of a model, we also want to evaluate its capability to localize an object.
For this, we evaluate the capabilities of our models to localize the center of the objects in an image by measuring the mean Average Euclidean Distance (mAED) between the ground truth and the closest estimated localization points. The details of the measurement are given in supplementary materials. To find the best configuration of matched points, we use the Hungarian algorithm\cite{Kuhn1955Hungarian}.

For the point-wise localization methods, we obtain directly the object center's $x,y$ coordinates. In detection, the localization coordinates are considered as the center of the detected objects. However, a classifier trained without explicit localization information does not output the objects' $x,y$ coordinates. Therefore, we employ Class Activation Maps \cite{DBLP:journals/corr/ZhouKLOT15} to extract the objects' center. Additional details regarding the algorithm used to find the positions from the activation maps can be found in the supplementary material.

\begin{figure}
    \includegraphics[width=\linewidth]{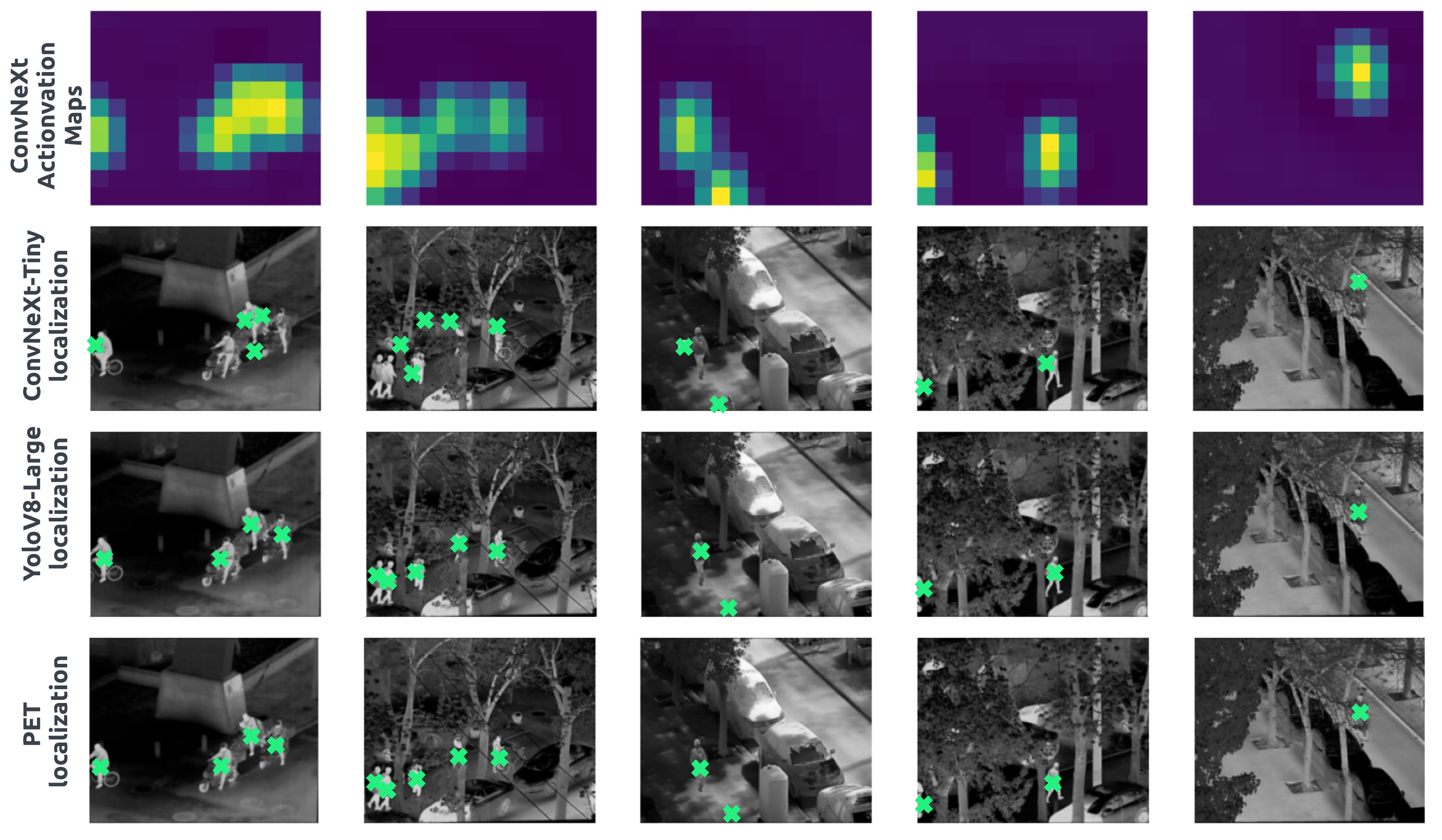}
    \caption{Examples of localizations of ConvNeXt, YoloV8 and PET}
    \label{fig:localiaztion}
\end{figure}

In \cref{tab:local}, we present the mAED results for our most promising models: YoloV8-L for detection, PET for point-wise localization, and ConvNeXt for image-level counting.
Despite the lack of location information in the labeling, ConvNeXt-Tiny achieved comparable results to YoloV8-L and outperformed PET. ConvNeXt localization excels when individuals are widely separated (\cref{fig:localiaztion}). However, the localization becomes imprecise when multiple people close, as evidenced in the first two images from ConvNeXt-Tiny. Such an issue is related to the low resolution of the activation maps, which limits the algorithm to delineate each maximum within closely clustered points accurately.

\begin{table}[!htp]\centering
    \caption{Mean average Euclidian Distance between ground truth points and estimated localization points for our algorithms with different level of supervision}\label{tab:local}
    \footnotesize
    \begin{tabular}{lcc}\toprule
        \textbf{Model} & \textbf{Supervision} & \textbf{mAED} \\\midrule
        YoloV8-L       & Stronger             & 0.12582       \\
        PET            & Strong               & 0.17054       \\
        ConvNeXt       & Normal               & 0.16857       \\
        \bottomrule
    \end{tabular}
\end{table}

\section{Conclusions}
\label{sec:concl}
In this study, we present key insights into people counting. Our analysis highlights several significant findings. First, regression heads for classifiers enhance performance, especially when dealing with balanced people counts, emphasizing the importance of data distribution in model outcomes. In addition, the Masked Autoencoder pretraining technique demonstrates its adaptability, proving effective not only in identifying object classes but also inaccurate people counting. Thirdly, our experiments reveal limitations in top-performing crowd counting methods based on point-level localization, particularly in sparse crowd scenarios, suggesting the need for further exploration in optimizing their efficiency. Moreover, this research underscores the feasibility of employing cost-effective annotation methods without compromising count accuracy, making people counting more accessible and affordable. Lastly, the study challenges the common belief that more supervision inherently translates to better results, as point-level methodologies under-performed compared to level supervision classifiers, underscoring the importance of aligning model design with crowd density for optimal outcomes.

\textbf{Acknowledgements:} This work was supported by Distech Controls Inc., NSERC, and MITACS. The authors thanks Heitor Rapela Medeiros and Masih Aminbeidokhti for the valuable discussion.

{\small
    \bibliographystyle{ieee_fullname}
    \bibliography{egbib}

\begin{thebibliography}{10}\itemsep=-1pt

\bibitem{DBLP:journals/corr/abs-2102-06171}
Andrew Brock, Soham De, Samuel~L. Smith, and Karen Simonyan.
\newblock High-performance large-scale image recognition without normalization.
\newblock {\em CoRR}, abs/2102.06171, 2021.

\bibitem{cai2023reversible}
Yuxuan Cai, Yizhuang Zhou, Qi Han, Jianjian Sun, Xiangwen Kong, Jun Li, and
  Xiangyu Zhang.
\newblock Reversible column networks, 2023.

\bibitem{10.1007/978-3-030-58452-8_13}
Nicolas Carion, Francisco Massa, Gabriel Synnaeve, Nicolas Usunier, Alexander
  Kirillov, and Sergey Zagoruyko.
\newblock End-to-end object detection with transformers.
\newblock In Andrea Vedaldi, Horst Bischof, Thomas Brox, and Jan-Michael Frahm,
  editors, {\em Computer Vision -- ECCV 2020}, pages 213--229, Cham, 2020.
  Springer International Publishing.

\bibitem{DBLP:journals/corr/abs-1908-03314}
Zhuojun Chen, Junhao Cheng, Yuchen Yuan, Dongping Liao, Yizhou Li, and
  Jiancheng Lv.
\newblock Deep density-aware count regressor.
\newblock {\em CoRR}, abs/1908.03314, 2019.

\bibitem{Cheng_2022_CVPR}
Zhi-Qi Cheng, Qi Dai, Hong Li, Jingkuan Song, Xiao Wu, and Alexander~G.
  Hauptmann.
\newblock Rethinking spatial invariance of convolutional networks for object
  counting.
\newblock In {\em Proceedings of the IEEE/CVF Conference on Computer Vision and
  Pattern Recognition}, pages 19638--19648, June 2022.

\bibitem{DBLP:journals/corr/abs-2010-11929}
Alexey Dosovitskiy, Lucas Beyer, Alexander Kolesnikov, Dirk Weissenborn,
  Xiaohua Zhai, Thomas Unterthiner, Mostafa Dehghani, Matthias Minderer, Georg
  Heigold, Sylvain Gelly, Jakob Uszkoreit, and Neil Houlsby.
\newblock An image is worth 16x16 words: Transformers for image recognition at
  scale.
\newblock {\em CoRR}, abs/2010.11929, 2020.

\bibitem{dubail2022privacypreserving}
Thomas Dubail, Fidel Alejandro~Guerrero Peña, Heitor~Rapela Medeiros, Masih
  Aminbeidokhti, Eric Granger, and Marco Pedersoli.
\newblock Privacy-preserving person detection using low-resolution infrared
  cameras, 2022.

\bibitem{FAN2022224}
Zizhu Fan, Hong Zhang, Zheng Zhang, Guangming Lu, Yudong Zhang, and Yaowei
  Wang.
\newblock A survey of crowd counting and density estimation based on
  convolutional neural network.
\newblock {\em Neurocomputing}, 472:224--251, 2022.

\bibitem{6460719}
Luca Fiaschi, Ullrich Koethe, Rahul Nair, and Fred~A. Hamprecht.
\newblock Learning to count with regression forest and structured labels.
\newblock In {\em Proceedings of the 21st International Conference on Pattern
  Recognition (ICPR2012)}, pages 2685--2688, 2012.

\bibitem{DBLP:journals/corr/Girshick15}
Ross~B. Girshick.
\newblock Fast {R-CNN}.
\newblock {\em CoRR}, abs/1504.08083, 2015.

\bibitem{DBLP:journals/corr/GirshickDDM13}
Ross~B. Girshick, Jeff Donahue, Trevor Darrell, and Jitendra Malik.
\newblock Rich feature hierarchies for accurate object detection and semantic
  segmentation.
\newblock {\em CoRR}, abs/1311.2524, 2013.

\bibitem{HABASH20221}
Riadh Habash.
\newblock 1 - building as a system.
\newblock In Riadh Habash, editor, {\em Sustainability and Health in
  Intelligent Buildings}, Woodhead Publishing Series in Civil and Structural
  Engineering, pages 1--32. Woodhead Publishing, 2022.

\bibitem{hastings1947low}
Cecil Hastings~Jr, Frederick Mosteller, John~W Tukey, and Charles~P Winsor.
\newblock Low moments for small samples: a comparative study of order
  statistics.
\newblock {\em The Annals of Mathematical Statistics}, 18(3):413--426, 1947.

\bibitem{DBLP:journals/corr/abs-2111-06377}
Kaiming He, Xinlei Chen, Saining Xie, Yanghao Li, Piotr Doll{\'{a}}r, and
  Ross~B. Girshick.
\newblock Masked autoencoders are scalable vision learners.
\newblock {\em CoRR}, abs/2111.06377, 2021.

\bibitem{DBLP:journals/corr/HeZRS15}
Kaiming He, Xiangyu Zhang, Shaoqing Ren, and Jian Sun.
\newblock Deep residual learning for image recognition.
\newblock {\em CoRR}, abs/1512.03385, 2015.

\bibitem{DBLP:journals/corr/abs-2108-10831}
Xinyu Jia, Chuang Zhu, Minzhen Li, Wenqi Tang, and Wenli Zhou.
\newblock {LLVIP:} {A} visible-infrared paired dataset for low-light vision.
\newblock {\em CoRR}, abs/2108.10831, 2021.

\bibitem{Kuhn1955Hungarian}
Harold~W. Kuhn.
\newblock {The Hungarian Method for the Assignment Problem}.
\newblock {\em Naval Research Logistics Quarterly}, 2(1--2):83--97, March 1955.

\bibitem{Li_2022_CVPR}
Feng Li, Hao Zhang, Shilong Liu, Jian Guo, Lionel~M. Ni, and Lei Zhang.
\newblock Dn-detr: Accelerate detr training by introducing query denoising.
\newblock In {\em Proceedings of the IEEE/CVF Conference on Computer Vision and
  Pattern Recognition (CVPR)}, pages 13619--13627, June 2022.

\bibitem{Liu_2023_ICCV}
Chengxin Liu, Hao Lu, Zhiguo Cao, and Tongliang Liu.
\newblock Point-query quadtree for crowd counting, localization, and more.
\newblock In {\em Proceedings of the IEEE/CVF International Conference on
  Computer Vision (ICCV)}, pages 1676--1685, October 2023.

\bibitem{liu2022dabdetr}
Shilong Liu, Feng Li, Hao Zhang, Xiao Yang, Xianbiao Qi, Hang Su, Jun Zhu, and
  Lei Zhang.
\newblock {DAB}-{DETR}: Dynamic anchor boxes are better queries for {DETR}.
\newblock In {\em International Conference on Learning Representations}, 2022.

\bibitem{DBLP:journals/corr/abs-1811-10452}
Weizhe Liu, Mathieu Salzmann, and Pascal Fua.
\newblock Context-aware crowd counting.
\newblock {\em CoRR}, abs/1811.10452, 2018.

\bibitem{DBLP:journals/corr/abs-2103-14030}
Ze Liu, Yutong Lin, Yue Cao, Han Hu, Yixuan Wei, Zheng Zhang, Stephen Lin, and
  Baining Guo.
\newblock Swin transformer: Hierarchical vision transformer using shifted
  windows.
\newblock {\em CoRR}, abs/2103.14030, 2021.

\bibitem{Liu_2021_ICCV}
Ze Liu, Yutong Lin, Yue Cao, Han Hu, Yixuan Wei, Zheng Zhang, Stephen Lin, and
  Baining Guo.
\newblock Swin transformer: Hierarchical vision transformer using shifted
  windows.
\newblock In {\em Proceedings of the IEEE/CVF International Conference on
  Computer Vision (ICCV)}, pages 10012--10022, October 2021.

\bibitem{DBLP:journals/corr/abs-2201-03545}
Zhuang Liu, Hanzi Mao, Chao{-}Yuan Wu, Christoph Feichtenhofer, Trevor Darrell,
  and Saining Xie.
\newblock A convnet for the 2020s.
\newblock {\em CoRR}, abs/2201.03545, 2022.

\bibitem{loshchilov2018decoupled}
Ilya Loshchilov and Frank Hutter.
\newblock Decoupled weight decay regularization.
\newblock In {\em International Conference on Learning Representations}, 2019.

\bibitem{inbook}
Chen~Change Loy, Ke Chen, Shaogang Gong, and Tao Xiang.
\newblock {\em Crowd Counting and Profiling: Methodology and Evaluation},
  volume~11.
\newblock 10 2013.

\bibitem{lu2022bridging}
Zhiying Lu, Hongtao Xie, Chuanbin Liu, and Yongdong Zhang.
\newblock Bridging the gap between vision transformers and convolutional neural
  networks on small datasets, 2022.

\bibitem{Ma_2022}
Yiming Ma, Victor Sanchez, and Tanaya Guha.
\newblock Fusioncount: Efficient crowd counting via multiscale feature fusion.
\newblock In {\em 2022 {IEEE} International Conference on Image Processing
  ({ICIP})}. {IEEE}, oct 2022.

\bibitem{7410729}
Viet-Quoc Pham, Tatsuo Kozakaya, Osamu Yamaguchi, and Ryuzo Okada.
\newblock Count forest: Co-voting uncertain number of targets using random
  forest for crowd density estimation.
\newblock In {\em 2015 IEEE International Conference on Computer Vision
  (ICCV)}, pages 3253--3261, 2015.

\bibitem{DBLP:journals/corr/RedmonDGF15}
Joseph Redmon, Santosh~Kumar Divvala, Ross~B. Girshick, and Ali Farhadi.
\newblock You only look once: Unified, real-time object detection.
\newblock {\em CoRR}, abs/1506.02640, 2015.

\bibitem{DBLP:journals/corr/RedmonF16}
Joseph Redmon and Ali Farhadi.
\newblock {YOLO9000:} better, faster, stronger.
\newblock {\em CoRR}, abs/1612.08242, 2016.

\bibitem{DBLP:journals/corr/abs-1804-02767}
Joseph Redmon and Ali Farhadi.
\newblock Yolov3: An incremental improvement.
\newblock {\em CoRR}, abs/1804.02767, 2018.

\bibitem{Ren17}
Peiming Ren, Wei Fang, and Soufiene Djahel.
\newblock A novel yolo-based real-time people counting approach.
\newblock 09 2017.

\bibitem{DBLP:journals/corr/RenHG015}
Shaoqing Ren, Kaiming He, Ross~B. Girshick, and Jian Sun.
\newblock Faster {R-CNN:} towards real-time object detection with region
  proposal networks.
\newblock {\em CoRR}, abs/1506.01497, 2015.

\bibitem{DBLP:journals/corr/RhodesQM16}
Anthony~D. Rhodes, Max~H. Quinn, and Melanie Mitchell.
\newblock Fast on-line kernel density estimation for active object
  localization.
\newblock {\em CoRR}, abs/1611.05369, 2016.

\bibitem{DBLP:journals/corr/SimonyanZ14a}
Karen Simonyan and Andrew Zisserman.
\newblock Very deep convolutional networks for large-scale image recognition.
\newblock In Yoshua Bengio and Yann LeCun, editors, {\em 3rd International
  Conference on Learning Representations, {ICLR} 2015, San Diego, CA, USA, May
  7-9, 2015, Conference Track Proceedings}, 2015.

\bibitem{song2021rethinking}
Qingyu Song, Changan Wang, Zhengkai Jiang, Yabiao Wang, Ying Tai, Chengjie
  Wang, Jilin Li, Feiyue Huang, and Yang Wu.
\newblock Rethinking counting and localization in crowds: A purely point-based
  framework.
\newblock 2021.

\bibitem{sun2018fishnet}
Shuyang Sun, Jiangmiao Pang, Jianping Shi, Shuai Yi, and Wanli Ouyang.
\newblock Fishnet: A versatile backbone for image, region, and pixel level
  prediction.
\newblock In {\em Advances in Neural Information Processing Systems}, pages
  760--770, 2018.

\bibitem{Tran_2022_BMVC}
Nguyen~Hoang Tran, Ta~Duc Huy, Soan T.~M. Duong, Phan Nguyen, Dao~Huu Hung,
  Chanh D~Tr Nguyen, Trung Bui, and QUOC~HUNG TRUONG.
\newblock Improving local features with relevant spatial information by vision
  transformer for crowd counting.
\newblock In {\em 33rd British Machine Vision Conference 2022, {BMVC} 2022,
  London, UK, November 21-24, 2022}. {BMVA} Press, 2022.

\bibitem{DBLP:journals/corr/abs-2009-13077}
Boyu Wang, Huidong Liu, Dimitris Samaras, and Minh Hoai.
\newblock Distribution matching for crowd counting.
\newblock {\em CoRR}, abs/2009.13077, 2020.

\bibitem{wang22crowds}
Q. Wang and T.P. Breckon.
\newblock Crowd counting via segmentation guided attention networks and
  curriculum loss.
\newblock {\em IEEE Trans. Intelligent Transportation Systems}, 2022.
\newblock to appear.

\bibitem{DBLP:journals/corr/abs-1912-12162}
Shuai Wang and Zhendong Su.
\newblock Metamorphic testing for object detection systems.
\newblock {\em CoRR}, abs/1912.12162, 2019.

\bibitem{DBLP:journals/corr/abs-2007-12831}
Yi Wang, Junhui Hou, Xinyu Hou, and Lap{-}Pui Chau.
\newblock A self-training approach for point-supervised object detection and
  counting in crowds.
\newblock {\em CoRR}, abs/2007.12831, 2020.

\bibitem{7533041}
Yi Wang and Yuexian Zou.
\newblock Fast visual object counting via example-based density estimation.
\newblock In {\em 2016 IEEE International Conference on Image Processing
  (ICIP)}, pages 3653--3657, 2016.

\bibitem{woo2023convnext}
Sanghyun Woo, Shoubhik Debnath, Ronghang Hu, Xinlei Chen, Zhuang Liu, In~So
  Kweon, and Saining Xie.
\newblock Convnext v2: Co-designing and scaling convnets with masked
  autoencoders, 2023.

\bibitem{DBLP:journals/corr/XieGDTH16}
Saining Xie, Ross~B. Girshick, Piotr Doll{\'{a}}r, Zhuowen Tu, and Kaiming He.
\newblock Aggregated residual transformations for deep neural networks.
\newblock {\em CoRR}, abs/1611.05431, 2016.

\bibitem{DBLP:journals/corr/abs-2106-13112}
Li Yuan, Qibin Hou, Zihang Jiang, Jiashi Feng, and Shuicheng Yan.
\newblock {VOLO:} vision outlooker for visual recognition.
\newblock {\em CoRR}, abs/2106.13112, 2021.

\bibitem{DBLP:journals/corr/abs-2106-11342}
Aston Zhang, Zachary~C. Lipton, Mu Li, and Alexander~J. Smola.
\newblock Dive into deep learning.
\newblock {\em CoRR}, abs/2106.11342, 2021.

\bibitem{zhang2023dino}
Hao Zhang, Feng Li, Shilong Liu, Lei Zhang, Hang Su, Jun Zhu, Lionel Ni, and
  Heung-Yeung Shum.
\newblock {DINO}: {DETR} with improved denoising anchor boxes for end-to-end
  object detection.
\newblock In {\em The Eleventh International Conference on Learning
  Representations}, 2023.

\bibitem{7780439}
Yingying Zhang, Desen Zhou, Siqin Chen, Shenghua Gao, and Yi Ma.
\newblock Single-image crowd counting via multi-column convolutional neural
  network.
\newblock In {\em 2016 IEEE Conference on Computer Vision and Pattern
  Recognition (CVPR)}, pages 589--597, 2016.

\bibitem{DBLP:journals/corr/ZhouKLOT15}
Bolei Zhou, Aditya Khosla, {\`{A}}gata Lapedriza, Aude Oliva, and Antonio
  Torralba.
\newblock Learning deep features for discriminative localization.
\newblock {\em CoRR}, abs/1512.04150, 2015.

\end{thebibliography}
}

\appendix

\section*{Supplementary Materials}

\section{Hyper-parameters}
The training hyper-parameters utilized for training the people counting models using image-level architectures, point-wise localization, and object detectors are outlined in \cref{tab:imagelevelparams}, \cref{tab:pointwiseparams}, and \cref{tab:detectorsparams}, respectively. These hyper-parameters were applied consistently across all reported results in Section 4. In the case of image-level tasks, the loss functions employed were either Mean Square Error (MSE) for regression or Cross Entropy (CE) for classification. On the other hand, point-wise localization loss functions encompassed multiple terms within the loss function, such as Euclidean distance between points (EUC), Smooth L1 distance (SL1), or Split loss (L-S). For object detectors, YoloV8 utilized Varifocal loss (VFL) and Distribution Focal loss, while DINO employed L1 distance and Generalized Intersection over Union (GIOU).

\cref{fig:threshcurves} displays the count accuracy versus threshold curves for object detectors. These curves were generated after the completion of training, and the validation set was utilized to identify the optimal score threshold for evaluation purposes. The determined best thresholds for the LLVIP and Distech IR datasets are outlined in \cref{tab:thresholds}.

\begin{table}[!htp]\centering
    \caption{Best thresholds per model per dataset}\label{tab:thresholds}
    \scriptsize
    \setlength{\tabcolsep}{4pt}
    \begin{tabular}{ccccc}\toprule
        \textbf{Dataset}    & \textbf{Yolov8-S} & \textbf{Yolov8-M} & \textbf{Yolov8-L} & \textbf{DINO} \\\midrule
        \textbf{LLVIP}      & 0.343             & 0.503             & 0.43              & 0.405         \\
        \textbf{Distech IR} & 0.351             & 0.159             & 0.504             & 0.405         \\
        \bottomrule
    \end{tabular}
\end{table}

\begin{table*}[!htp]
    \centering
    \caption{Image-level training hyper-parameters}\label{tab:imagelevelparams}
    \scriptsize
    \begin{tabular}{lrrrrrrr}\toprule
                                         & \multicolumn{2}{c}{\textbf{Image-Level models from Scratch}} & \multicolumn{2}{c}{\textbf{Image-Level models from MAE Pretraining}} & \multicolumn{2}{c}{\textbf{Image-Level models from Fine-Tuning}}                                                    \\\cmidrule{2-7}
                                         & ConvNeXt-Micro/Tiny                                          & ViT-3L/4L                                                            & ConvNeXt-Micro/Tiny                                              & ViT-3L/4L   & ConvNeXt-Micro/Tiny & ViT-3L/4L    \\\midrule
        \textbf{Learning Rate}           & 1.00e-4                                                      & 1.00e-4                                                              & 1.50e-6                                                          & 1.50e-4     & 2.00e-4             & 1.00e-5      \\
        \textbf{Epochs}                  & 450                                                          & 450                                                                  & 500                                                              & 500         & 450                 & 450          \\
        \textbf{Batch Size}              & 64                                                           & 64                                                                   & 64                                                               & 64          & 64                  & 64           \\
        \textbf{Optimizer}               & AdamW                                                        & AdamW                                                                & AdamW                                                            & AdamW       & AdamW               & AdamW        \\
        \textbf{Momentum (beta1, beta2)} & (0.9; 0.999)                                                 & (0.9; 0.95)                                                          & (0.9; 0.95)                                                      & (0.9; 0.95) & (0.9; 0.999)        & (0.9; 0.999) \\
        \textbf{Weight Decay}            & 0.3                                                          & 0.3                                                                  & 0.05                                                             & 0.05        & 0.05                & 0.3          \\
        \textbf{Scheduler}               & Cosine                                                       & Cosine                                                               & Cosine                                                           & Cosine      & Cosine              & Cosine       \\
        \textbf{Warmup Epochs}           & 40                                                           & 20                                                                   & 40                                                               & 40          & 40                  & 5            \\
        \textbf{Loss Function}           & MSE/CE                                                       & MSE/CE                                                               & MSE                                                              & MSE         & MSE/CE              & MSE/CE       \\
        \bottomrule
    \end{tabular}
\end{table*}

\begin{table*}[!htp]
    \parbox{.5\linewidth}{
        \centering
        \caption{Point-wise localization training hyper-parameters}\label{tab:pointwiseparams}
        \scriptsize
        \begin{tabular}{lrrr}\toprule
                                             & \multicolumn{2}{c}{\textbf{Point-Level Localizers}}                  \\\cmidrule{2-3}
                                             & P2PNet                                              & PET            \\\midrule
            \textbf{Learning Rate}           & 1.00e-4                                             & 1.00e-4        \\
            \textbf{Epochs}                  & 1500                                                & 1500           \\
            \textbf{Batch Size}              & 8                                                   & 8              \\
            \textbf{Optimizer}               & Adam                                                & AdamW          \\
            \textbf{Momentum (beta1, beta2)} & (0.9, 0.999)                                        & (0.9, 0.999)   \\
            \textbf{Weight Decay}            & N/A                                                 & N/A            \\
            \textbf{Scheduler}               & None                                                & None           \\
            \textbf{Warmup Epochs}           & N/A                                                 & N/A            \\
            \textbf{Loss Function}           & EUC + CE                                            & SL1 + CE + L-S \\
            \bottomrule
        \end{tabular}
    }
    \parbox{.45\linewidth}{
        \centering
        \caption{Object Detection training hyper-parameters}\label{tab:detectorsparams}
        \scriptsize
        \begin{tabular}{lrrr}\toprule
                                                       & \multicolumn{2}{c}{\textbf{Object Detectors}}                  \\\cmidrule{2-3}
                                                       & YoloV8-S/M/L                                  & DINO-SWIN-Tiny \\\midrule
            \textbf{Learning Rate}                     & 1.00e-3                                       & 1.00e-4        \\
            \textbf{Epochs}                            & 400                                           & 12             \\
            \textbf{Batch Size}                        & 16                                            & 4              \\
            \textbf{Optimizer}                         & SGD                                           & AdamW          \\
            \textbf{Momentum [beta or (beta1, beta2)]} & 0.937                                         & (0.9; 0.999)   \\
            \textbf{Weight Decay}                      & 0.0005                                        & N/A            \\
            \textbf{Scheduler}                         & Cosine                                        & None           \\
            \textbf{Warmup Epochs}                     & 3                                             & N/A            \\
            \textbf{Loss Function}                     & VFL + DFL                                     & L1 + GIOU      \\
            \bottomrule
        \end{tabular}
    }
\end{table*}

\begin{figure*}
    \centering
    \setlength{\tabcolsep}{0pt}
    \begin{tabular}{cccc}
        \multicolumn{4}{c}{LLVIP dataset}                                                                                                                                                                                                                                                                                                                                                                             \\
        \includegraphics[width=.25\linewidth]{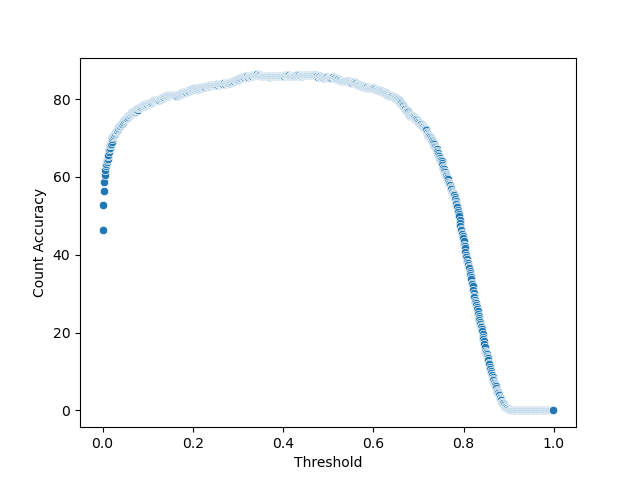}                     & \includegraphics[width=.25\linewidth]{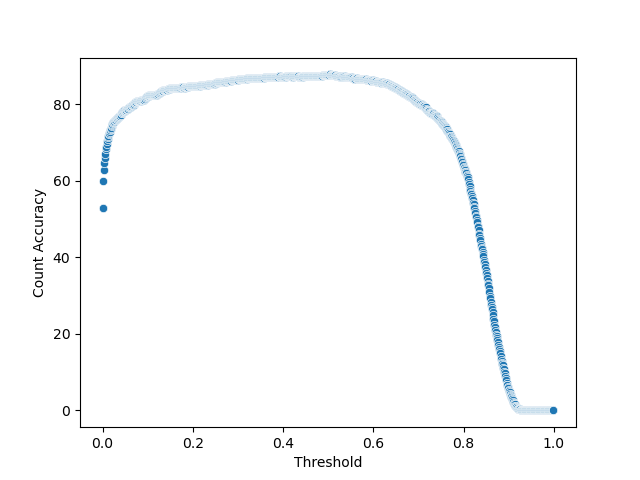}                     & \includegraphics[width=.25\linewidth]{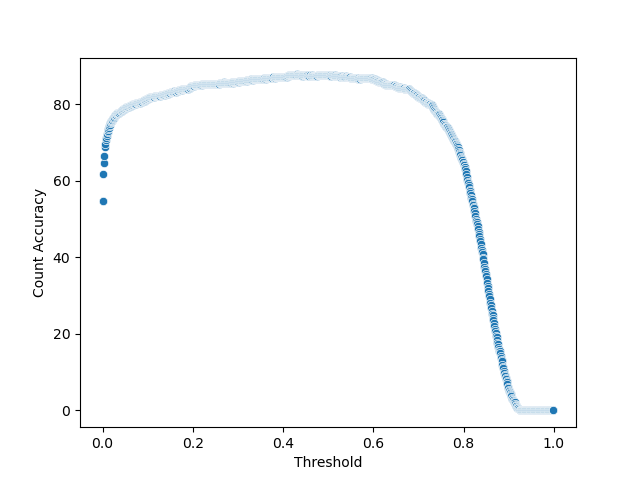}                     &
        \includegraphics[width=.25\linewidth]{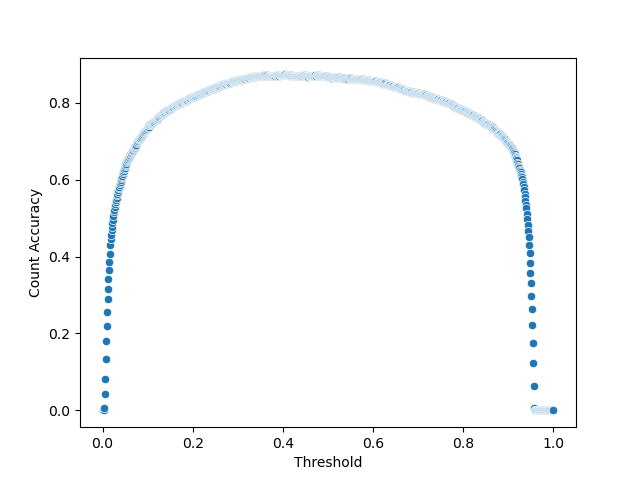}                                                                                                                                                                                                                                                                                                                         \\
        \multicolumn{4}{c}{Distech IR dataset}                                                                                                                                                                                                                                                                                                                                                                        \\
        \includegraphics[width=.25\linewidth]{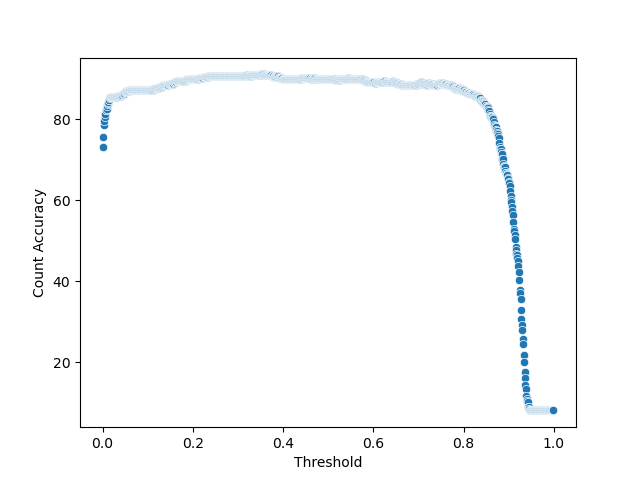} & \includegraphics[width=.25\linewidth]{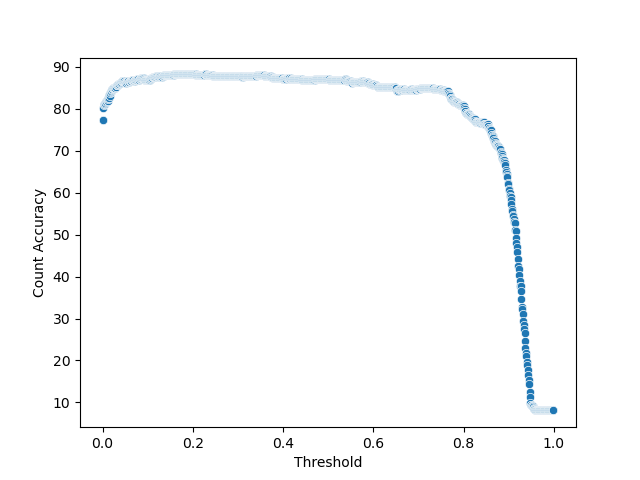} & \includegraphics[width=.25\linewidth]{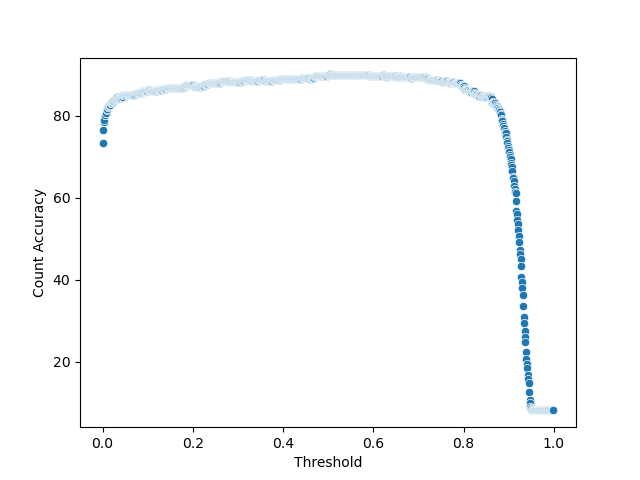} & \includegraphics[width=.25\linewidth]{materials/LLVIP_DINO_threshold_tuning_accs.png} \\
        YoloV8-S                                                                                              & YoloV8-M                                                                                              & YoloV8-L                                                                                              & DINO                                                                                  \\
    \end{tabular}

    \caption{Count accuracy at various thresholds for the LLVIP and Distech IR datasets using YoloV8-S, YoloV8-M, YoloV8-L, and DINO models. The selected best threshold represents the value that yields the highest count accuracy.}
    \label{fig:threshcurves}
\end{figure*}

\section{Effect of class imbalance on image-level counting}
\cref{tab:cillvip} and \cref{tab:cidistech} illustrate the achieved count accuracy per number of people in each image for LLVIP and Distech IR, respectively. The tables also present the class distribution, aiding in the analysis of class imbalance within this framework. As expected, higher occurrences of a category correlate with higher count accuracy. Classes represented by only one image displayed a binary performance outcome, typically 0\% or 100\%. This trend aligns with findings in existing literature on image-level tasks characterized by severe class imbalance. Notably, regression-based methods were equally impacted by this class imbalance. These results are a notable drawback of such people counting techniques. We hypothesize that employing stratified versions of datasets or techniques designed to mitigate the effects of class imbalance might enhance performance and potentially match the performance of object detectors.

\begin{table*}[!htp]\centering
    \caption{Count Accuracy Per Class LLVIP}\label{tab:cillvip}
    \scriptsize
    \setlength{\tabcolsep}{4pt}
    \begin{tabular}{lrrrrrrrrrrrrrrrrr}\toprule
                                  &                       & \textbf{Accuracy Count : } & 0                     & 1     & 2     & 3     & 4     & 5     & 6     & 7     & 8     & 9     & 10    & 11   & 12     & 13     \\\cmidrule{3-17}
                                  &                       & \textbf{Occurences}        & 1                     & 1203  & 990   & 602   & 333   & 171   & 81    & 45    & 24    & 9     & 3     & 2    & 1      & 1      \\\cmidrule{3-17}
        \textbf{Model}            & \textbf{Pretrain.}    & \textbf{Head Type}         & \multicolumn{14}{c}{}                                                                                                          \\\midrule
        \multirow{4}{*}{ConvNeXt} & \multirow{2}{*}{None} & Classification             & 100.00                & 91.81 & 83.52 & 74.61 & 73.32 & 65.99 & 48.78 & 57.50 & 29.41 & 7.14  & 25.00 & 0.00 & 100.00 & 0.00   \\
                                  &                       & Regression                 & 100.00                & 92.47 & 85.86 & 74.61 & 68.27 & 69.23 & 53.66 & 66.25 & 44.12 & 21.43 & 37.50 & 0.00 & 100.00 & 100.00 \\
                                  & \multirow{2}{*}{MAE}  & Classification             & 100.00                & 92.79 & 86.16 & 73.05 & 70.91 & 67.61 & 57.72 & 60.00 & 35.29 & 7.14  & 0.00  & 0.00 & 100.00 & 100.00 \\
                                  &                       & Regression                 & 100.00                & 93.56 & 85.15 & 75.70 & 71.15 & 69.23 & 52.03 & 56.25 & 38.24 & 28.57 & 25.00 & 0.00 & 100.00 & 100.00 \\
        \multirow{4}{*}{ViT}      & \multirow{2}{*}{None} & Classification             & 0.00                  & 80.79 & 65.41 & 54.98 & 56.01 & 42.51 & 38.21 & 36.25 & 26.47 & 0.00  & 25.00 & 0.00 & 100.00 & 0.00   \\
                                  &                       & Regression                 & 0.00                  & 78.82 & 66.73 & 58.26 & 55.53 & 48.99 & 34.96 & 35.00 & 20.59 & 0.00  & 0.00  & 0.00 & 0.00   & 0.00   \\
                                  & \multirow{2}{*}{MAE}  & Classification             & 0.00                  & 86.79 & 64.90 & 55.30 & 57.21 & 36.84 & 26.83 & 50.00 & 2.94  & 0.00  & 0.00  & 0.00 & 0.00   & 0.00   \\
                                  &                       & Regression                 & 100.00                & 81.66 & 63.58 & 61.99 & 56.25 & 45.34 & 39.84 & 42.50 & 29.41 & 14.29 & 12.50 & 0.00 & 0.00   & 100.00 \\
        \bottomrule
    \end{tabular}
\end{table*}

\begin{table*}[!htp]\centering
    \caption{Count Accuracy Per Class  Distech IR}\label{tab:cidistech}
    \scriptsize
    \begin{tabular}{lrrrrrrrrr}\toprule
                                  &                       & \textbf{Accuracy Count : } & 0     & 2     & 3     & 4     & 5     & 6      \\\cmidrule{3-9}
                                  &                       & \textbf{Occurences}        & 19    & 50    & 36    & 5     & 144   & 1      \\\cmidrule{3-9}
        \textbf{Model}            & \textbf{Pretraining}  & \textbf{Head Type}         &       &       &       &       &       &        \\\midrule
        \multirow{4}{*}{ConvNeXt} & \multirow{2}{*}{None} & Classification             & 89.47 & 86.00 & 44.44 & 20.00 & 95.83 & 0.00   \\
                                  &                       & Regression                 & 94.74 & 62.00 & 83.33 & 20.00 & 93.75 & 0.00   \\
                                  & \multirow{2}{*}{MAE}  & Classification             & 84.21 & 82.00 & 75.00 & 0.00  & 97.22 & 100.00 \\
                                  &                       & Regression                 & 89.47 & 78.00 & 77.78 & 20.00 & 93.75 & 0.00   \\
        \multirow{4}{*}{ViT}      & \multirow{2}{*}{None} & Classification             & 94.74 & 68.00 & 30.56 & 40.00 & 96.53 & 0.00   \\
                                  &                       & Regression                 & 94.74 & 56.00 & 52.78 & 0.00  & 86.81 & 0.00   \\
                                  & \multirow{2}{*}{MAE}  & Classification             & 89.47 & 76.00 & 33.33 & 0.00  & 91.67 & 100.00 \\
                                  &                       & Regression                 & 78.95 & 42.00 & 33.33 & 20.00 & 84.72 & 0.00   \\
        \bottomrule
    \end{tabular}
\end{table*}

\section{Localization details}

The localization results on the main manuscript are reported in terms of mean Absolute Euclidean Distance (mAED). We calculate the mAED between the predicted coordinates and the ground truth for all images in the testing set. Both the $x$ and $y$ position coordinates are normalized within the range of 0 to 1.

The mAED is computed as:
\begin{equation}
    mAED =  \frac{1}{M} \sum_{j=1}^{M}  \frac{1}{N_j} \sum_{i=1}^{N_j} ||p_{ij}-\hat{p}_{ij}||_{2}^{2}
    \label{eq:approx}
\end{equation}
where $\hat{p}_{ij}$ and $p_{ij}$ represent the coordinates of the predicted and ground truth positions for the $i^{th}$ point within the $j^{th}$ image. As previously outlined, the predicted points and ground truth are paired using the Hungarian matching algorithm. In this context, $M$ signifies the total number of images within the test set, while $N_j$ indicates the total number of points per image. When the estimated points and ground truth locations fail to align, a penalty distance of 1 is assigned.

\begin{figure*}
    \begin{tabular}{c}
        \includegraphics[width=18cm]{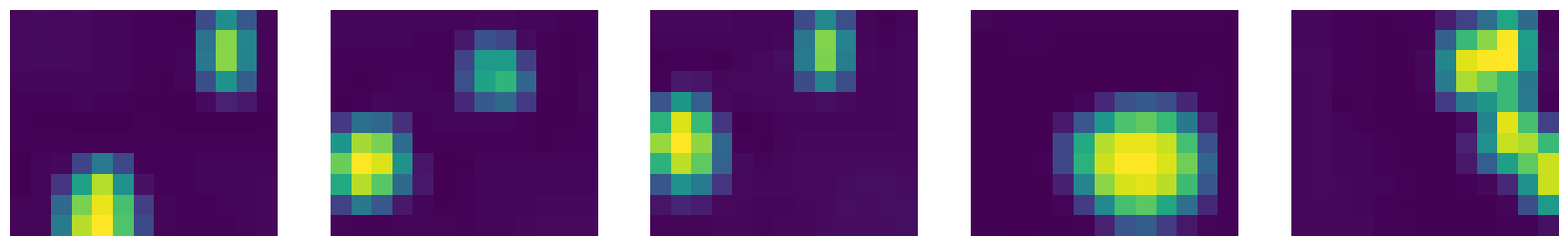} \\
        \includegraphics[width=18cm]{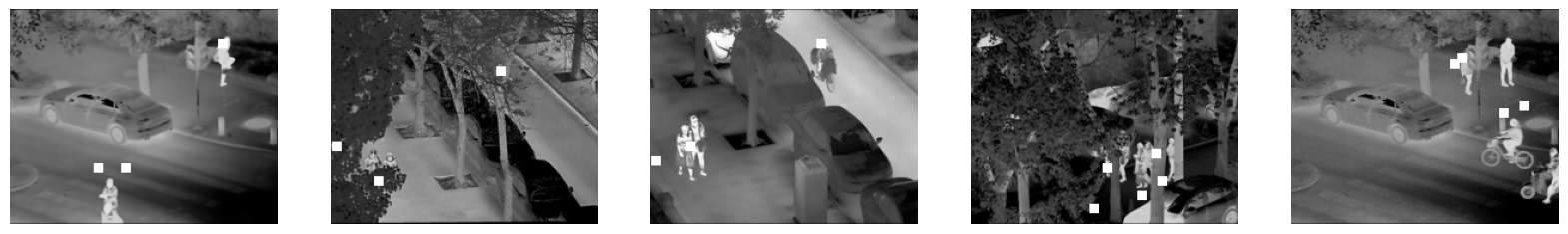} \\
        \includegraphics[width=18cm]{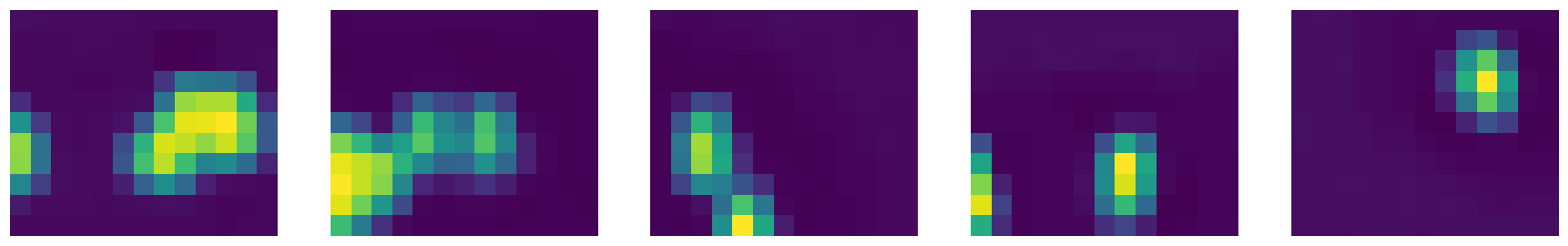} \\
        \includegraphics[width=18cm]{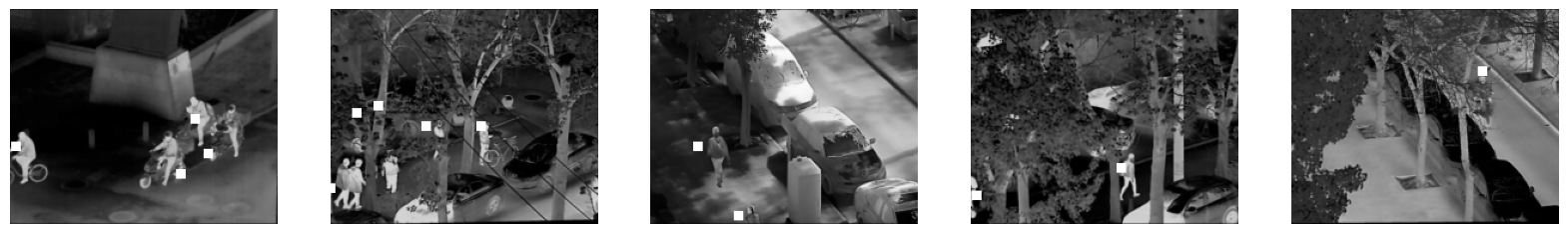} \\
        \includegraphics[width=18cm]{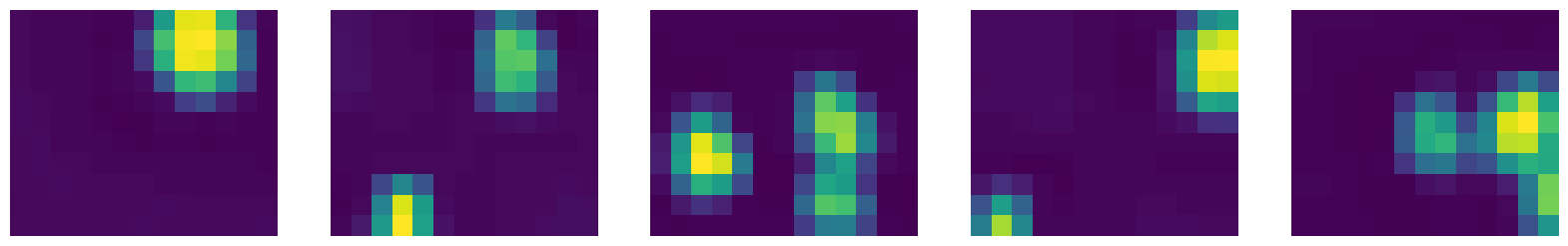} \\
        \includegraphics[width=18cm]{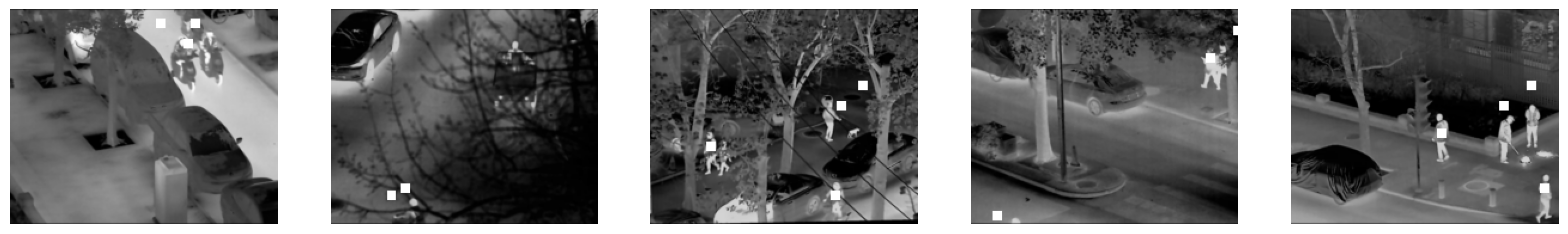} \\
        \includegraphics[width=18cm]{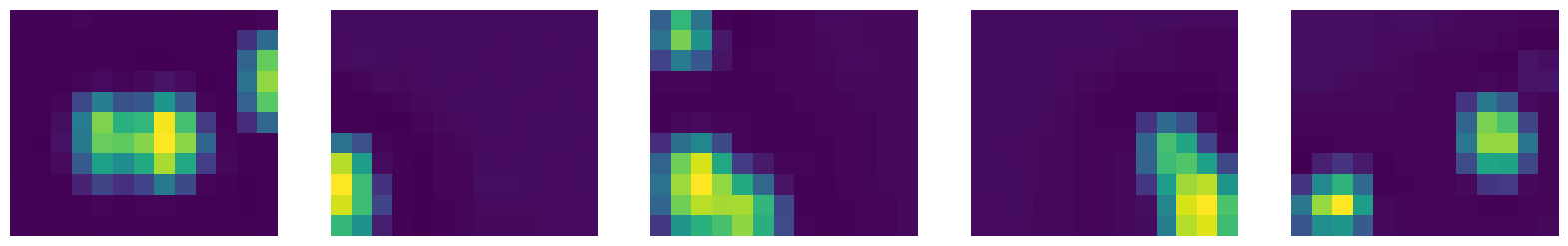} \\
        \includegraphics[width=18cm]{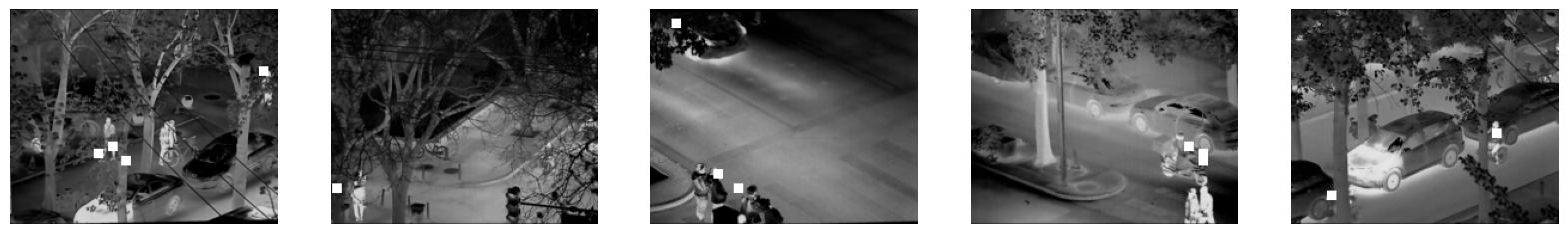} \\
    \end{tabular}
    \caption{Examples of people localization using ConvNeXt attention maps.}
    \label{fig:local}
\end{figure*}

Since image-level techniques lack localization information, we use class activation maps to obtain the locations of individuals. Our proposed algorithm leverages the identified count of people ($numInstances$) to guide te localization process. Empirically, a threshold of 27 was determined to effectively binarize the activation map and extract the relevant regions of interest. For specifics regarding the employed algorithm, refer to \cref{algo:euclid}. Several examples illustrating the localization obtained using the aforementioned algorithm from the activation maps are presented in \cref{fig:local}.

\begin{algorithm*}
    \caption{People's location from ConvNeXt activation maps}\label{algo:euclid}
    \begin{algorithmic}[1]
        \Procedure{LocatePeople}{activationMap, binaryThreshold, numInstances}
        \State $activationMap \gets \textbf{binarize}(activationMap,binaryThreshold)$
        \State $countours \gets \textbf{findCountours}(activationMap)$.
        \State $x,y \gets \textbf{findCoordinates}(countours)$.
        \If{$numInstances = \textbf{len}(countours)$}
        \State \Return $x, y$
        \ElsIf{$numInstances < \textbf{len}(countours)$}
        \State $ countours \gets \textbf{reverseSortByAreaSize}(countours)$
        \State $x, y \gets \textbf{findCoordinates}(countours)$.
        \State $x, y \gets \textbf{sliceList}(countours, start=0, end=numInstances)$.
        \State \Return $x, y$
        \Else
        \State $x, y \gets List(), List()$
        \State $avgInstanceSize \gets \textbf{totalArea}(countours) / numInstances$
        \For{$contour \text{ in } countours$}
        \State $numPeoples \gets \textbf{round}(\textbf{area}(contour) / avgInstanceSize)$
        \If{$numPeoples \leq 1$}
        \State $px,py \gets \textbf{findCoordinates}(countour)$
        \State $x \gets x.\textbf{append}(px)$
        \State $y \gets y.\textbf{append}(py)$
        \Else
        \State $randomCoordinates \gets \textbf{uniformRandomSample}(contour, \text{numSamples}=numPeoples)$
        \For{$px, py \text{ in } randomCoordinates$}
        \State $x \gets x.\textbf{append}(px)$
        \State $y \gets y.\textbf{append}(py)$
        \EndFor
        \EndIf
        \EndFor
        \State \Return $x, y$
        \EndIf
        \EndProcedure
    \end{algorithmic}
\end{algorithm*}

\end{document}